\documentclass[sigconf]{acmart}
\newcommand{\textBC}[2]{\textbf{\textcolor{#1}{#2}}}
\usepackage{amssymb}
\usepackage{epsfig}
\usepackage{graphicx}
\usepackage{tabularx}
\usepackage{tabu}
\usepackage{amsmath}
\usepackage{subfigure}
\usepackage{booktabs}
\usepackage{multirow}
\usepackage{makecell,rotating}
\usepackage{indentfirst}
\usepackage{enumitem}
\usepackage{color}
\usepackage{url}
\AtBeginDocument{%
  \providecommand\BibTeX{{%
    \normalfont B\kern-0.5em{\scshape i\kern-0.25em b}\kern-0.8em\TeX}}}

\copyrightyear{2021} 
\acmYear{2021} 
\setcopyright{acmcopyright}
\acmConference[MM '21]{Proceedings of the 29th ACM International Conference on Multimedia}{October 20--24, 2021}{Virtual Event, China}\acmBooktitle{Proceedings of the 29th ACM International Conference on Multimedia (MM '21), October 20--24, 2021, Virtual Event, China}
\acmPrice{15.00}
\acmDOI{10.1145/3474085.3475192}
\acmISBN{978-1-4503-8651-7/21/10}

\copyrightyear{2021} 
\acmYear{2021} 
\setcopyright{acmcopyright}\acmConference[MM '21]{Proceedings of the 29th ACM
	International Conference on Multimedia}{October 20--24, 2021}{Virtual Event, China}
\acmBooktitle{Proceedings of the 29th ACM International Conference on Multimedia
	(MM '21), October 20--24, 2021, Virtual Event, China}
\acmPrice{15.00}
\acmDOI{10.1145/3474085.3475192}
\acmISBN{978-1-4503-8651-7/21/10}

\begin{document}
\fancyhead{}
\title{Multi-Source Fusion and Automatic Predictor Selection \\ for Zero-Shot Video Object Segmentation}


\author{Xiaoqi Zhao}
\email{zxq@mail.dlut.edu.com}
\affiliation{%
  \institution{Dalian University of Technology}
  \streetaddress{No.2 Linggong Road, Ganjingzi District}
  \city{Dalian}
  \state{Liaoning}
  \country{China}
  \postcode{116024}
}
\author{Youwei Pang}
\email{lartpang@mail.dlut.edu.com}
\affiliation{%
  \institution{Dalian University of Technology}
  \streetaddress{No.2 Linggong Road, Ganjingzi District}
  \city{Dalian}
  \state{Liaoning}
  \country{China}
  \postcode{116024}
}
\author{Jiaxing Yang}
\email{jx.yang@mail.dlut.edu.com}
\affiliation{%
  \institution{Dalian University of Technology}
  \streetaddress{No.2 Linggong Road, Ganjingzi District}
  \city{Dalian}
  \state{Liaoning}
  \country{China}
  \postcode{116024}
}

\author{Lihe Zhang}
\authornote{Corresponding author.}
\email{zhanglihe@dlut.edu.com}
\authornotemark[0]
\affiliation{%
  \institution{Dalian University of Technology}
  \streetaddress{No.2 Linggong Road, Ganjingzi District}
  \city{Dalian}
  \state{Liaoning}
  \country{China}
  \postcode{116024}
}
\author{Huchuan Lu}
\email{lhchuan@dlut.edu.com}
\affiliation{%
  \institution{Dalian University of Technology}
  \streetaddress{No.2 Linggong Road, Ganjingzi District}
  \city{Dalian}
  \state{Liaoning}
  \country{China}\\
  \institution{Pengcheng Lab}
   \city{Shenzhen}
  \state{Guangdong}
  \country{China}
  \postcode{116024}
}



\begin{abstract}
  Location and appearance are the key cues for video object segmentation. Many sources such as RGB, depth, optical flow and static saliency can provide useful information about the objects. However, existing approaches only utilize the RGB or RGB and optical flow. 
In this paper, we propose a novel multi-source fusion network for zero-shot video object segmentation. With the help of interoceptive spatial attention module (ISAM), spatial importance  of each source is highlighted. Furthermore, we design a feature purification module (FPM) to filter the inter-source incompatible features. By the ISAM and FPM, the multi-source features are effectively fused.
In addition, we put forward an automatic predictor selection network (APS) to select the better prediction of either the static saliency predictor or the moving object predictor in order to prevent over-reliance on the failed results caused by low-quality optical flow maps. 
Extensive experiments on three challenging public benchmarks (i.e. DAVIS$_{16}$, Youtube-Objects and FBMS) show that the proposed model achieves compelling performance against the state-of-the-arts. The source code will be publicly available at \textcolor{red}{\url{https://github.com/Xiaoqi-Zhao-DLUT/Multi-Source-APS-ZVOS}}.
\end{abstract}




\begin{teaserfigure}
  \includegraphics[width=\textwidth]{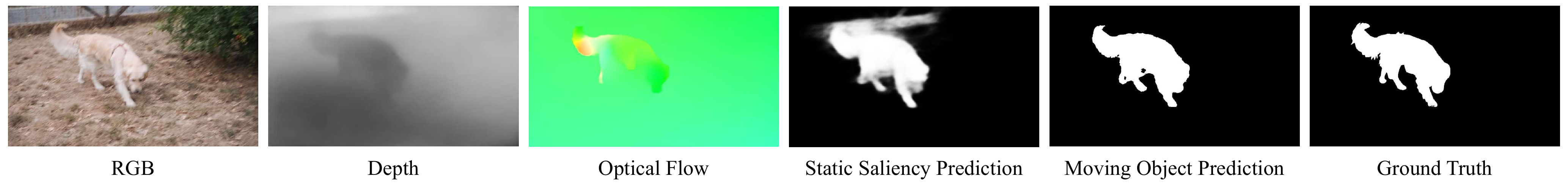}
  \caption{Visual results of different sources. }
  \label{fig:Figure1}
\end{teaserfigure}
\maketitle

\section{Introduction}
Zero-shot video object segmentation (ZVOS) aims to automatically separate primary foreground object/objects from their background in a video without any human annotation for testing frames. It has attracted a lot of interests due to the wide application scenarios such as autonomous driving, video surveillance and video editing.
With the development of deep convolutional neural networks (CNNs), the CNNs-Based
ZVOS methods dominate this field. According to the way of capturing the moving objects, ZVOS methods can be  divided into  interframe-based~\cite{WCS,AGNN,COSNet,AGS,EPO,PDB} and optical flow based methods~\cite{MATNet,GateNet,MP,SFL}. In this paper, we utilize the optical flow to depict motion information of video objects. 

To achieve accurate video object segmentation, the input information sources are important. As we know, a video sequence is made up of a series of static images. The process of observing objects is from static to dynamic. If the object in a video no longer moves or moves very slowly, its segmentation is converted to a static salient object segmentation problem. 
When there is obvious relative motion in the scene, the optical flow map contains the  patterns of objects, surfaces and edges. 
In addition, the depth map can also provide useful complementary information for segmentation tasks, such as temporal action localization~\cite{jiyuan_mm}, RGB-D semantic ~\cite{PA-RGBD-SS,DA-RGBD-SS,SS-RGBD-SS} and  saliency segmentation~\cite{ACM_RGBD_SOD_1,DANet,HDFNet}. 
Thus, the RGB, optical flow, depth and static saliency all can provide the vital position and appearance cues about video objects, and each source has the complementarity, as shown in Fig.~\ref{fig:Figure1}. 

\begin{figure}[t]
\includegraphics[width=\linewidth]{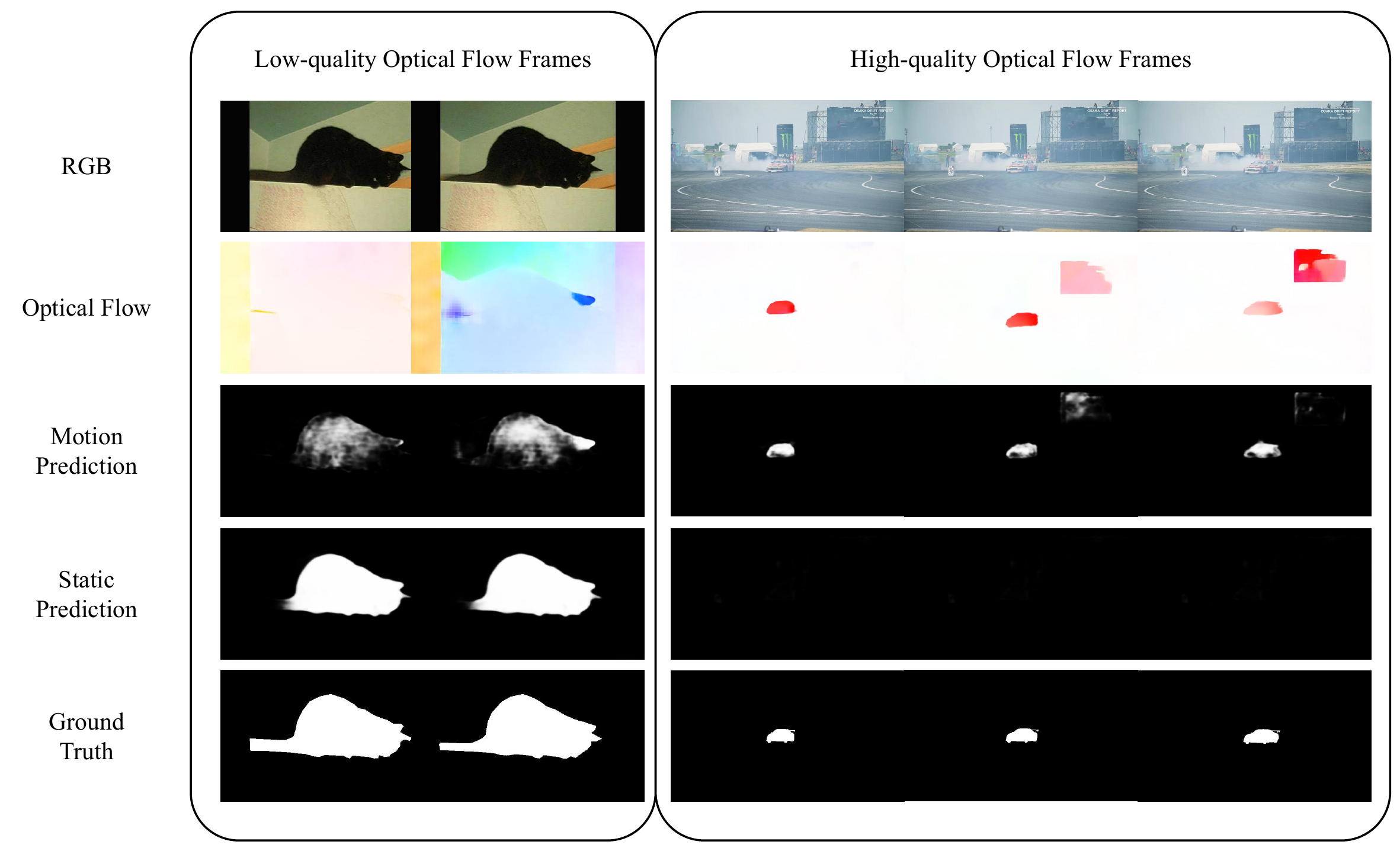}\\
        \centering
        \caption{Visual results of static and motion prediction.}
\label{fig:Figure2}
\vspace{-5mm}
\end{figure}

However, all previous ZVOS methods only focus on the RGB or RGB and optical flow, other sources are neglected. Besides, existing optical flow based approaches extremely rely on the optical flow at multiple levels. If the foreground (object) shifts significantly, the optical flow map can capture the object well, which is beneficial to the network. On the contrary, if the background changes drastically or the foreground hardly moves, the resulted optical flow may be noise for ZVOS. As shown in Fig.~\ref{fig:Figure2}, the high-quality optical flow can provide effective guidance, while the low-quality one easily brings the interference. How to avoid this problem is not considered by previous optical flow based methods.

Motivated by these observations, we propose a novel multi-source fusion and automatic predictor selection network. 
Firstly, we design a simple multi-task network, which aims to predict the depth map and static saliency map from RGB image. It adopts the FPN~\cite{FPN} structure with one encoder and two decoders. The encoder extracts the RGB features, while the decoders infer the depth and static saliency features, respectively.
Secondly, we design an interoceptive spatial attention module (ISAM) to effectively combine the feature maps provided by four kinds of sources (i.e, RGB, depth, optical flow and static saliency). 
The ISAM can adaptively perceive the importance of each source features in their spatial positions compared to other sources, thereby preserving the source-specific information in the fused features. 
Since multiple sources  contain some mutual interference effects, we build a feature purification module (FPM) 
to filter out the incompatible information. With the help of ISAM and FPM, the moving object can be segmented precisely.
Lastly, we design an automatic predictor selection network (APS) to evaluate the objectness of the optical flow and choose the result between static salient object segmentation and moving
object segmentation, thereby avoiding the prediction failure caused by the optical flow.

Our main contributions can be summarized as follows:
\begin{itemize}
     \item  We present a novel solution and new insight for  zero-shot video object segmentation by utilizing multiple sources to provide comprehensive location and appearance information of video objects.
     
    \item  We design an interoceptive spatial attention module (ISAM) and a feature purification module (FPM) to obtain multi-source compatible features.
   
     \item We propose an  automatic predictor selection network to adaptively choose better prediction from either the static saliency predictor or the moving object predictor. 
    
    \item Experimental results indicate that the proposed method significantly surpasses the existing state-of-the-art algorithms on three popular ZVOS benchmarks DAVIS$_{16}$, Youtube-Objects and FBMS.
\end{itemize}

\begin{figure*}
    \includegraphics[width=\textwidth]{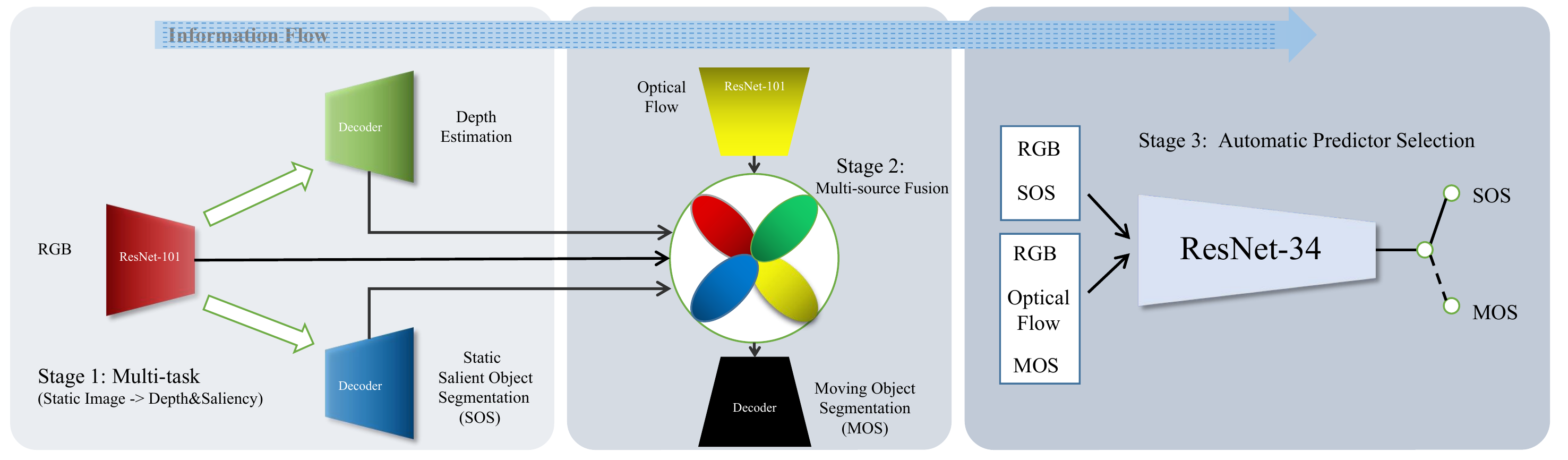}\\ 
    \centering
    \caption{Network pipeline of the ZVOS task. It consists of three stages: multi-task prediction, multi-source fusion and automatic predictor selection. The first stage network is used to generate features of RGB, depth and static saliency. The interoceptive spatial attention module (ISAM) and the feature purification module (FPM) are equipped in the second stage network to achieve multi-source fusion. The third stage network selects a better prediction from either the static object predictor or moving object predictor as the final output.} 		
    \label{fig:Figure3}
\end{figure*} 

\section{Related Work}
\subsection{Zero-shot Video Object Segmentation}
Different from one-shot video object segmentation (OVOS), ZVOS aims to automatically detect the target object without any human definition. Many CNNs-based methods~\cite{PDB, MotAdapt, EPO, AGS, COSNet, AGNN} are proposed to utilize the inter-frame relationship to capture rich context and enable a more complete understanding of video content. For instance, recurrent neural network is used to capture temporal information in~\cite{PDB,AGS}. Lu~\textit{et al.}~\cite{COSNet} take a pair of frames as input and learn their correlations by using co-attention mechanism. Wang~\textit{et al.}~\cite{AGNN} propose an attended graph neural network and perform recursive message passing to mine the underlying high-order correlations.

In addition, optical flow itself can provide important motion information. Benefiting from some outstanding optical flow estimation methods~\cite{RAFT,MaskFlowNet,PWC}, optical flow map can be easily obtained and applied to ZVOS.
Tokmakov~\textit{et al.}~\cite{MP} only use the optical flow map as the input and build a fully convolutional network to segment the moving object. But this map can not provide sufficient appearance information compared to the RGB input. In~\cite{LVO,SFL,UVOS-Bilateral,MATNet}, two parallel streams are built to extract features from the RGB image and optical flow map, which are further fused in the decoder to predict the segmentation results. The MATNet~\cite{MATNet} achieves the transition of attended motion features to enhance appearance learning at each convolution stage. However, the aforementioned optical flow based methods depend on the optical flow map heavily. 
This map deeply participates in the feature fusion of the network. 
Once its quality is very low, it is bound to result in very serious interference. To address this issue, we put forward an automatic predictor selection network to judge the effectiveness of the optical flow based predictor.

\subsection{Multi-source Information}
The RGB refers to three channels of red, green and blue. This standard includes almost all colors that human vision can perceive, and is one of the most widely used color systems. Many computer vision tasks, such as classification, object detection, object tracking and semantic segmentation all use the RGB image as the main input source. However, only relying on the RGB source is difficult to handle some complex environments such as low-contrast objects, which share similar appearances to the background. 
Benefiting from  Microsoft  Kinect and  Intel  RealSense devices, depth information can be conveniently obtained. Moreover, the stable geometric structures depicted in the depth map are  robust  against the changes of illumination and texture, which can provide important supplementary information for handling complex scenarios. 
Therefore, many RGB-D methods are applied in different tasks, such as RGB-D semantic segmentation,  RGB-D salient object segmentation, 
and RGB-D tracking~\cite{rgbd-tracking1,rgbd-tracking2,rgbd-tracking3}. 

In addition, as an important computer vision task, salient object segmentation also needs to delineate the location and contour information of salient objects in a scene, which can provide crucial cues for ZVOS.
Many methods~\cite{GateNet,BASNet,DSS} are proposed to solve this basic vision task. 
For the video related task, motion information is a key attribute. 
How to effectively capture motion prior has received much attention, where optical flow estimation is an active research branch. 
Recently, many CNNs-based methods~\cite{SFN,PWC,liteflownet,VCN} utilize iterative refinement strategy to improve the performance of optical flow.
%
In this work, we aim to exploit the aforementioned multiple sources to solve the zero-shot video object segmentation.

\section{Methodology}
Overall, we segment objects of both motion and appearance saliency within video frames by employing a three-stage pipeline, as shown in Fig.~\ref{fig:Figure3}. In the first stage, a network capable of achieving simultaneously depth estimation and static salient object segmentation (SOS) is built, which only feeds on static images. 
In the second, a network able to fuse features from different sources (RGB image, depth, static saliency, and optical flow) is engineered to achieve moving object segmentation (MOS), with an interoceptive spatial attention module (ISAM) and a feature purification module (FPM) at its core. 
In the third, a discriminator network is applied to judge SOS and MOS outputs according to their respective confidence scores, to self-adaptively decide which is the desirable one. The following content will detail the specifics of the three stages in order and show how these specified networks are coordinated.
\begin{figure}[t]
\includegraphics[width=\linewidth]{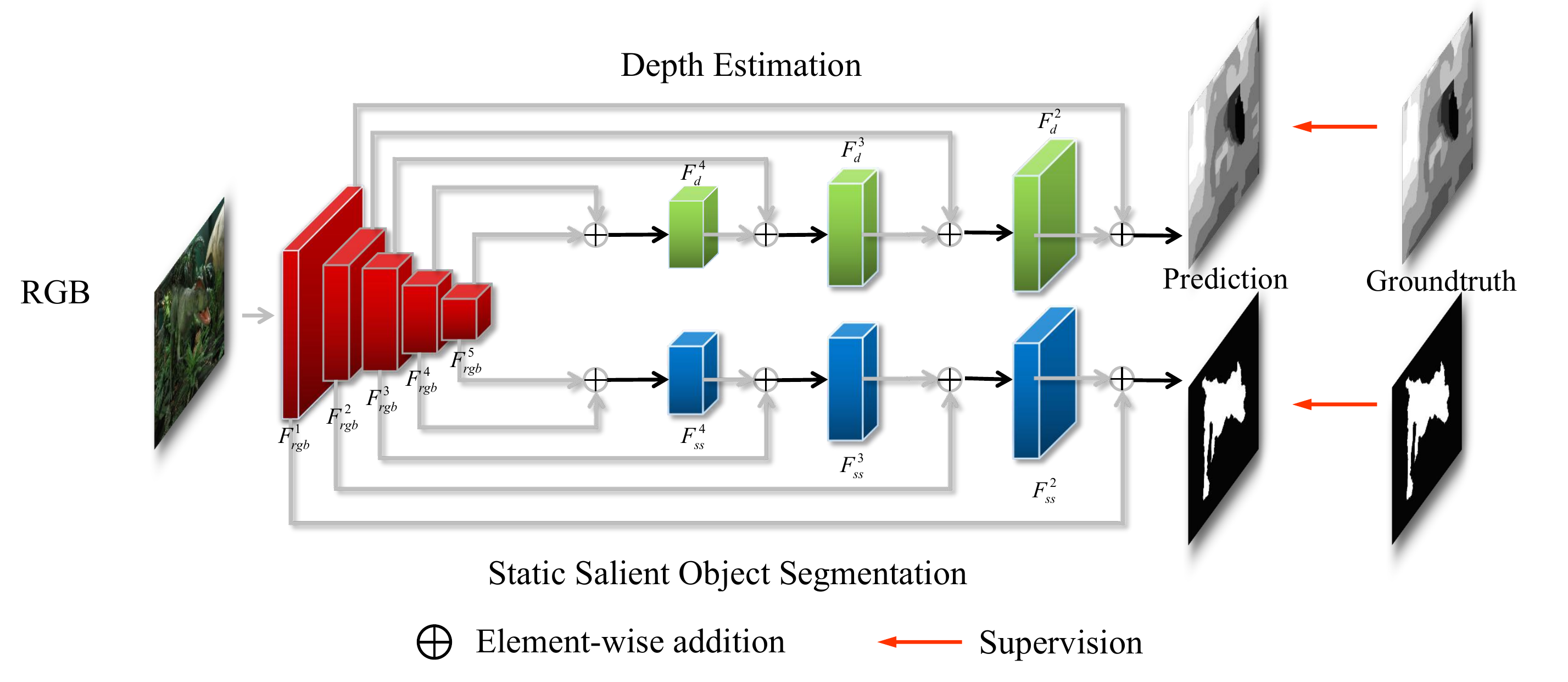}\\
        \centering
        \caption{Illustration of the Multi-task network.}
\label{fig:Figure4}
\vspace{-5mm}
\end{figure} 
\subsection{Multi-task Prediction}\label{sec:RGB-D source}
The multi-task network in this stage (adjusted from feature pyramid networks~\cite{FPN}) has a uniencoder-bidecoder structure, as illustrated in Fig.~\ref{fig:Figure4}, in which the encoder is fed with static images and the decoders will predict depth and static saliency maps, respectively. The encoder uses tail-cast ResNet-101~\cite{Resnet} to accommodate the requirement of our task (fully-convolutional). The two-stream decoders have five upsampling stages
to gradually restore resolution of the embedding, during which the features from each encoder stage are fused into their corresponding decoder stages via the self-explaining skip-connection. In addition, we apply foreground object ground truth and depth annotation to supervise their outputs, respectively. For depth stream, we follow the related works~\cite{depth3,depth4,depth5} to adopt the combination loss of L1 and SSIM~\cite{SSIM}. For another, we use BCE loss~\cite{BCE} as in ~\cite{DSS,Amulet,DGRL,GateNet,MINet}. Such double supervision strategy will drive the transformation of source semantics, namely, RGB source to depth source and static saliency source. Note that compared to the methods directly adopting existing depth estimation and saliency detection networks, our multi-task design will save training and inference time and a number of parameters.

\begin{figure*}
\includegraphics[width=\linewidth]{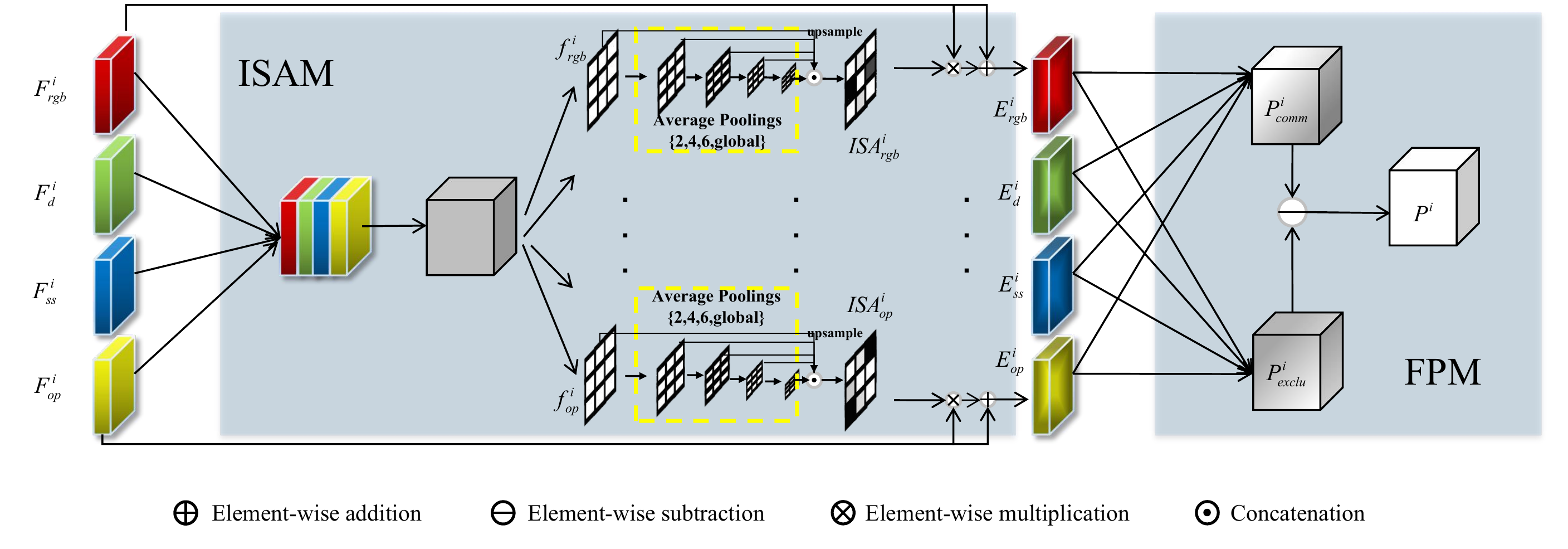}
        \centering
        \caption{Detailed diagram of interoceptive spatial attention module (ISAM) and feature purification module (FPM).}
\label{fig:Figure5}
\end{figure*} 
To be specific, our multi-task network will produce: (1) the RGB feature map $F_{rgb}^i$ ($i\in\{1, 2, 3, 4, 5\}$) which is extracted from the $i$th level of the encoder; (2) the depth feature map $F_{d}^i$  from the decoder of depth estimation; (3) the static saliency feature map $F_{ss}^i$ from the decoder of salient object segmentation; (4) the depth map and the static saliency map $M_{sos}$.  The above three kinds of feature maps will be fed into the second stage network together for moving object segmentation. 




\subsection{Multi-source Fusion}

The multi-source fusion network consists of ISAM, FPM, an encoder, and a decoder, as shown in Fig.~\ref{fig:Figure3}. The encoder and decoder are similar to their counterparts in the first stage. Note that the encoder in this stage is only used to extract motion feature maps $F_{op}^i$  from optical flow map, which is calculated by feeding temporal close frames into the off-the-shelf RAFT Net~\cite{RAFT}. Thus, the features of the four sources are well-prepared. The ISAM enhances the spatial saliency of the feature maps of the sources by computing four attention maps. The FPM calculates the difference between features containing inter-source common information and mutually-exclusive information to filter out incompatible contexts. Their internal structures are shown in Fig.~\ref{fig:Figure5}.  

Within the network, ISAM first concatenates all the source feature maps to generate a single-channel interoceptive feature map for each source, as follows:
\begin{equation}\label{equ:1}
    {f}_{src}^i = Conv_{1}(Conv_{256}(Cat(F_{rgb}^i, F_{d}^i, F_{op}^i, F_{ss}^i))),
\end{equation}
where $src\in\{rgb, d, op, ss\}$ describes the identities of RGB image source, depth source, optical flow source, and static saliency source, respectively. The $Conv_{256}(\cdot)$ and $Conv_{1}(\cdot)$ operations refer to the $3\times 3$ convolutions with $256$ output channels and one output channel, respectively, by which the independent source features are correlated.
$Cat(\cdot)$ is the concatenation operation along channel axis. 
Then, we compute an interoceptive attention map
with multi-scale information. This process is formulated as follows:
\begin{equation}\label{equ:2}
    {ISA}_{src}^i = Sig(Conv_{1}(Cat(Up(MP({f}_{src}^i))), 
\end{equation}
where $MP$ denotes a group of pooling operations along the channel dimension with scale $\{2, 4, 6, global\}$, $Up(\cdot)$ is the bilinear interpolation to upsample the features map to the same size as ${f}_{src}^{i}$ and $Sig(\cdot)$ is the element-wise sigmoid function.  The ${ISA}_{src}^i$ is used to enhance each source feature as follows: 
\begin{equation}\label{equ:3}
    \begin{split}
        {E}_{src}^i = {F}_{src}^i + {F}_{src}^i \otimes {ISA}_{src}^i, 
    \end{split}
\end{equation}
where $\otimes$ represents element-wise multiplication, and ${E}_{src}^i$ is the enhanced feature at the $i$th level.

Following ISAM, FPM at every level takes the concatenation of ${E}_{src}^i$ from different sources as inputs to obtain the fused feature. There is the incompatibility problem when fusing multi-source features. Direct combination ${P}_{comm}^i$ is not sufficient to dilute incompatible components. Therefore, we construct an auxiliary feature, namely ${P}_{exclu}^i$.
The fusion process is formulated as:
\begin{equation}\label{equ:4}
    \begin{split}
        {P}^{i} = {P}_{comm}^i - {P}_{exclu}^i, 
    \end{split}
\end{equation}
where ${P}_{comm}^i$ and ${P}_{exclu}^i$ can be represented as:  
\begin{equation}\label{equ:44}
    \begin{split}
       {P}_{comm}^i = Conv_{256}(Cat({E}_{rgb}^i, {E}_{d}^i, {E}_{op}^i, {E}_{ss}^i), \\
       {P}_{exclu}^i = Conv_{256}(Cat({E}_{rgb}^i, {E}_{d}^i, {E}_{op}^i, {E}_{ss}^i).
    \end{split}
\end{equation}
${P}_{comm}^i$ and ${P}_{exclu}^i$ have different convolution parameters although their formulas are the same. The rationale behind FPM is that the substraction and the MOS-used supervision will force features ${P}_{comm}^i$ and ${P}_{exclu}^i$ to represent the common and mutually-exclusive contexts of the four sources, respectively.  After preparing ${P}^{i}$ at each level, we gradually combine all of them to generate the final prediction $M_{mos}$ of moving object segmentation by the plain decoder.

\subsection{Automatic Predictor Selection}
\begin{figure*}
\includegraphics[width=0.7\linewidth]{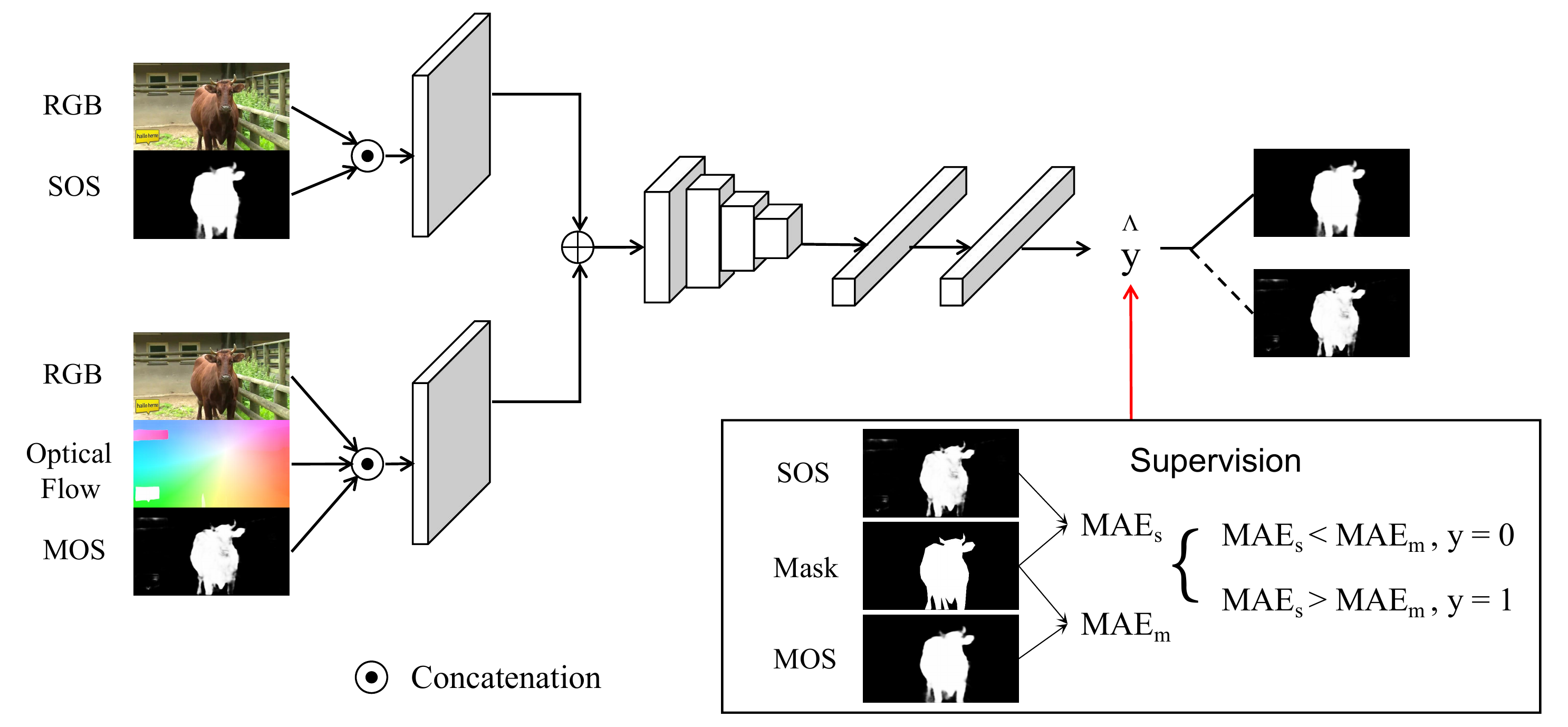}\\
        \centering
        \caption{The overall architecture of the automatic predictor selection network.}
\label{fig:Figure6}
\end{figure*}

In the third stage, the proposed automatic predictor selection network (APS) will judge which of the outputs from SOS and MOS is more convincing, according to a score. As shown in Fig.~\ref{fig:Figure6}, the overall network is constructed to solve a binary classification problem. We adopt a lightweight ResNet-34~\cite{Resnet} network in order to reduce the amount of parameters and also for easier training.
During the forward propagation, two inputs are required: (1) the concatenation of RGB and the SOS result; (2) the concatenation of RGB, optical flow map and the MOS result. For mathematical tractability, we use $rs$ and $rm$ to represent them, respectively. We replace the first layer of the encoder with the convolution of $64$ output channels, so that it can perceive the two inputs of different sizes. 
We initialize the parameters of the added convolutional layers with He's method~\cite{PRelu}.
After the first layer, we integrate the two-stream feature outputs by element-wise addition. The fused feature maps are fed into the other layers of ResNet-34 pretrained on ImageNet~\cite{ImageNet}, to alleviate the over-fitting problem to some extent. 
The process of the score prediction can be formulated as follows:
	\begin{equation}\label{equ:5}
	\begin{split}
	\hat y = APS(rs, rm;\theta),
	\end{split} 
	\end{equation}
where $\theta$ is the learnable parameters and $\hat y$ is the score. To sharp the discrimination ability of APS, we utilize the MAE score to dynamically generate the annotation $y$:
\begin{equation}\label{equ:6}
\centering
    \left\{\begin{matrix}
    \mathcal{M}(M_{sos}, G) < \mathcal{M}(M_{mos},G),  y = 0\\
    \mathcal{M}(M_{sos}, G) >= \mathcal{M}(M_{mos},G), y = 1,
    \end{matrix}\right.
\end{equation}
where $\mathcal{M}$ is the MAE defined as the average pixel-wise absolute difference between the prediction $M$ and ground-truth $G$:
	\begin{equation}\label{equ:7}
	\begin{split}
	MAE = \dfrac{1}{W \times H}\sum_{w=1}^W\sum_{h=1}^H(M(w,h)-G(w,h)),
	\end{split} 
	\end{equation}
where 
$W$, $H$ are the width
and height of the ground-truth mask, respectively. 
Finally, we adopt the binary cross entropy loss, which can be computed as:
 \begin{equation}\label{equ:8}
 \begin{split}
    \mathcal{L}_{bce} = -(ylog\hat {y}+(1-y)log(1-\hat{y})).
 \end{split}
\end{equation}

Supervised by the dynamic binary classification annotation, the APS network can learn the predictor selection rule that a lower score represents a higher confidence of the SOS and a lower degree of matching among MOS, optical flow and RGB. On the contrary, a higher score means a higher confidence of the MOS and a lower degree of matching between SOS and RGB. Therefore, the APS network implements the function of evaluating the effectiveness of object-information contained in the optical flow for ZVOS.

\section{Experiments}
\subsection{Datasets and Evaluation Metrics}

\emph{DAVIS$_{16}$}~\cite{davis16} is one of the most popular benchmark datasets for video object segmentation task. It consists of $50$ high-quality video sequences ($30$ for training and $20$ for validation) in total. And most videos do not have special scenes such as dramatic changes in the background or almost no movement in the foreground. Usually, the high-quality optical flow map can be obtained via optical flow networks. We use two standard metrics: mean similarity $\mathcal{J}$ and mean boundary accuracy $\mathcal{F}$~\cite{davis16} to measure the segmentation results (The higher are better). 

\emph{Youtube-Objects}~\cite{youtube-objects} is a large dataset of $126$ web videos with $10$ semantic object categories and more than $20,000$ frames. This dataset contains many unconventional videos, such as dramatically-changing backgrounds, slowly-moving or stationary objects, objects only moving in the depth dimension. The quality of optical flow map is usually low in these video sequences. Following most methods~\cite{MATNet,WCS,AGNN,COSNet,AGS,PDB}, we only use the mean region similarity $\mathcal{J}$ metric to measure the performance. 

\emph{FBMS}~\cite{FBMS} is composed of $59$ video sequences with ground truth annotations provided in a subset of the frames. Following the standard protocol~\cite{LVO} and other methods~\cite{MATNet,PDB}, we do not use any sequences within for training and only evaluate on the validation set, which consists of $30$ sequences. Since it is designed for their specific purposes, FBMS is not often used for the ZVOS task. We just use it for a fair comparison.
\subsection{Implementation Details}
Our model is implemented based on Pytorch and trained on a PC with an RTX 2080Ti GPU. The input sources are all resized to $384 \times 384$ and all the three stages use the  mini-batch of size $4$ for training.
\begin{table*}
	\small
	\centering
	\caption{Quantitative comparison on the DAVIS$_{16}$~\cite{davis16} validation set.
		The best result for each metric is \textBC{red}{red}.
		All the results are borrowed from the public leaderboard maintained by the DAVIS challenge or the corresponding papers.}
	\resizebox{0.9\textwidth}{!} {
	\begin{tabular}{cr||cccccccc||cccccc}
	   \toprule[1pt]
	    & & &\multicolumn{5}{c}{Interframe-based methods}&&&\multicolumn{5}{c}{Optical flow-based methods} \\ \hline
		& Methods  &PDB  &MotAdapt  & EPO &AGS & COSNet  & AGNN & DFNet & WCS & SFL  & MP& GateNet&MATNet &Ours \\ 
		& &~\cite{PDB}&~\cite{MotAdapt}&~\cite{EPO}&~\cite{AGS}&\cite{COSNet}&~\cite{AGNN}&~\cite{DFNet}&~\cite{WCS}&~\cite{SFL}&~\cite{MP}&~\cite{GateNet}&~\cite{MATNet}&ours\\
		\hline \hline
		& mean $\mathcal{J}$ $\uparrow$   &77.2  &77.2  &80.6  &79.7  &80.5  & 80.7& 80.4 &82.2 &67.4  & 70.0 & 80.9 &82.4 & \textBC{red}{83.3}   \\
		\hline
		& mean $\mathcal{F}$ $\uparrow$   &74.5  &77.4  &75.5  &77.4  &79.5  & 79.1 & -- & 80.7 & 66.7  &65.9  & 79.4  &80.7 & \textBC{red}{82.1} \\
 	 \bottomrule[1pt]
	\end{tabular}
	}
	\label{table:Table1}
\end{table*}
\begin{table*}
\Large
	\centering
	\caption{
	Quantitative results of each category on the Youtube-Objects~\cite{youtube-objects} in terms of mean $\mathcal{J}$. We show the average performance for each of the 10 categories, and the final row gives an average over all the videos.}
	\resizebox{0.9\textwidth}{!}
	{
		\setlength\tabcolsep{4pt}
		\renewcommand\arraystretch{1}
		\begin{tabular}{c||cccccccccc|c}
		 \toprule[1pt]
			& Airplane (6) & Bird (6) & Boat (15)  & Car (7) & Cat (16)& Cow (20) & Dog (27) &Horse (14)& Motorbike (10)& Train (5) & Avg. \\
			\hline
			\hline
			FST~\cite{FST}& 70.9 &70.6& 42.5&65.2  &52.1&44.5&65.3 &53.5&44.2 &29.6 &53.8  \\		
			COSEG~\cite{COSEG}&69.3   &76.0 &53.5  &70.4 & 66.8& 49.0 & 47.5& 55.7& 39.5& 53.4& 58.1\\
			ARP~\cite{ARP} &73.6 & 56.1&57.8 &33.9& 30.5& 41.8& 36.8 &44.3& 48.9& 39.2&46.2\\
			LVO~\cite{LVO} &86.2 &81.0&68.5 &69.3& 58.8&68.5&61.7 &53.9& 60.8& 66.3&67.5\\
			PDB~\cite{PDB} &78.0  &80.0 &58.9 &76.5 &63.0&64.1&70.1 &67.6&58.3&35.2&65.4\\
			FSEG~\cite{FSEG}  &{81.7}  &63.8 &{72.3}&74.9&68.4&68.0 &69.4 & 60.4&62.7&\textBC{red}{62.2}&68.4\\
		
AGS~\cite{AGS} &{87.7}&76.7&72.2&78.6&69.2&64.6&73.3&64.4&62.1&48.2&69.7\\
COSNet~\cite{COSNet} &81.1&75.7&71.3&77.6&66.5&{69.8}&76.8&67.4&67.7&46.8&70.5\\
AGNN~\cite{AGNN} &81.1  &75.9 &70.7 &78.1 &67.9&69.7&{77.4} &67.3&68.3&47.8&70.8\\
WCS~\cite{WCS} &81.8  &81.2 &67.6 &79.5 &65.8&66.2&73.4 &{69.5}&\textBC{red}{69.3}&49.7&70.9\\
\hline\hline
SFL~\cite{SFL}&65.6  &65.4 &59.9&64.0 &58.9&51.1& 54.1 &64.8& 52.6& 34.0&57.0\\
MATNet~\cite{MATNet} &72.9  &77.5 &66.9 &79.0 &\textBC{red}{73.7}&67.4&75.9 &63.2&62.6&51.0&69.0\\
Ours &\textBC{red}{89.4}  &\textBC{red}{83.7} &\textBC{red}{75.6} &\textBC{red}{80.7} &72.8 &\textBC{red}{71.6} &\textBC{red}{78.8}&\textBC{red}{70.7}&{63.1}&\textBC{red}{62.2}&\textBC{red}{74.9}\\ \bottomrule[1pt]
\end{tabular}
}
\label{table:Table2}
\end{table*}
First, we use some RGB-D saliency datasets~\cite{early_fusion_1,NJU2000,SIP,RGBD135,STERE,DMRA} to train the multi-task network. Once the first stage training is finished, we adopt the DAVIS$_{16}$ training set to train the multi-source fusion network. In the process, the parameters of the multi-task network is frozen, and we just train the multi-source fusion network. 
We use a total of $8,363$ labeled samples including video data ($2,000$ + frames) and 
\begin{table}[t]
	\small
	\centering
	\setlength{\tabcolsep}{0.3em}	
	 \setlength{\abovecaptionskip}{3pt}
	\caption{Quantitative results on the FBMS~\cite{FBMS} dataset.}
	\begin{tabular}{c||cccccccc}
	\hline
	Methods	& ARP& SAGE & LVO& FSEG & PDB & MATNet&Ours\\ 
	&~\cite{ARP}&~\cite{SAGE}&~\cite{LVO}&~\cite{FSEG}&~\cite{PDB}&~\cite{MATNet}&ours\\\hline
	mean $\mathcal{J}\uparrow$ & 59.8 & 61.2& 65.1& 68.4& 74.0& 76.1&\textBC{red}{76.7}\\
	\hline
	\end{tabular}
	\vspace{-5mm}
	\label{table:Table3}	
\end{table}
RGB-D SOD data ($6300$ + images) in the first and second phases. The depth map is obtained by the depth camera. MATNet~\cite{MATNet} uses video data ($14,000$ + frames)for training. AGS~\cite{AGS} uses video data ($8,500$ + frames) and RGB SOD data ($6,000$ + images) for training. COSNet~\cite{COSNet}, AGNN~\cite{AGNN} and GateNet~\cite{GateNet} use video data ($2,000$ + frames) and RGB SOD data ($15,000$ + images). In addition, we expect that there are low-quality and high-quality optical flow videos for training in the third stage, while almost all of the optical flow maps in the DAVIS$_{16}$ dataset are high-quality, it is not suitable for this training. Therefore, we use $4,000$ + frames from the DAVSOD dataset to train the APS. In the inference process, we generate the binary classification using the threshold value of $0.5$. In a word, the scale of the used annotations in our training phase is comparable to that of these competitors. 

We adopt some data augmentation techniques in each stage to avoid over-fitting: horizontally random flip, random rotate, random brightness, saturation and contrast. For the optimizers, we all use the SGD with a momentum of $0.9$ and a weight decay of $0.0005$. The learning rate is set to $0.005$ and later use the ``poly'' policy~\cite{poly} with the power of $0.9$ as a means of adjustment.

\subsection{Comparison with State-of-the-art}
\noindent\textbf{Performance on DAVIS$_{16}$.}
We compare the proposed method with the state-of-the-art ZVOS methods on the DAVIS$_{16}$ benchmark~\cite{davis16}. Tab.~\ref{table:Table1} shows performance comparison results in terms of the mean $\mathcal{J}$ and mean $\mathcal{F}$.
It can be seen that ours can consistently outperforms other approaches under the two metrics.  Compared to the second best \& the optical flow-based method (MATNet), our method achieves an important improvement of $1.1\%$ and $1.7\%$ in terms of mean $\mathcal{J}$ and mean $\mathcal{F}$, respectively. Moreover, we test the FLOPs of Ours vs. MATNet is 89.6G vs. 120.5G. Therefore, our method has significant advantages in terms of accuracy and computational complexity.\\  
\textbf{Performance on Youtube-Objects.} Tab.~\ref{table:Table2} shows the results on Youtube-Objects. It can be seen that our method achieves the best performance in eight of the ten categories.  Notably, our method has a significant performance improvement of $8.6\%$ in terms of mean $\mathcal{J}$ compared to the optical flow-based method MATNet ($74.9$ vs. $69.0$). As mentioned by Zhou \textit{et al.}~\cite{MATNet}, Youtube-Objects has many video sequences containing slowly moving and/or visually indistinct objects. Both will result in inaccurate estimation of optical flow. Therefore, the existing optical flow based methods can not perform very well on this dataset. With the help of our automatic predictor selection network, this problem can be solved well.\\
\begin{figure*}
    \includegraphics[width=0.8\textwidth]{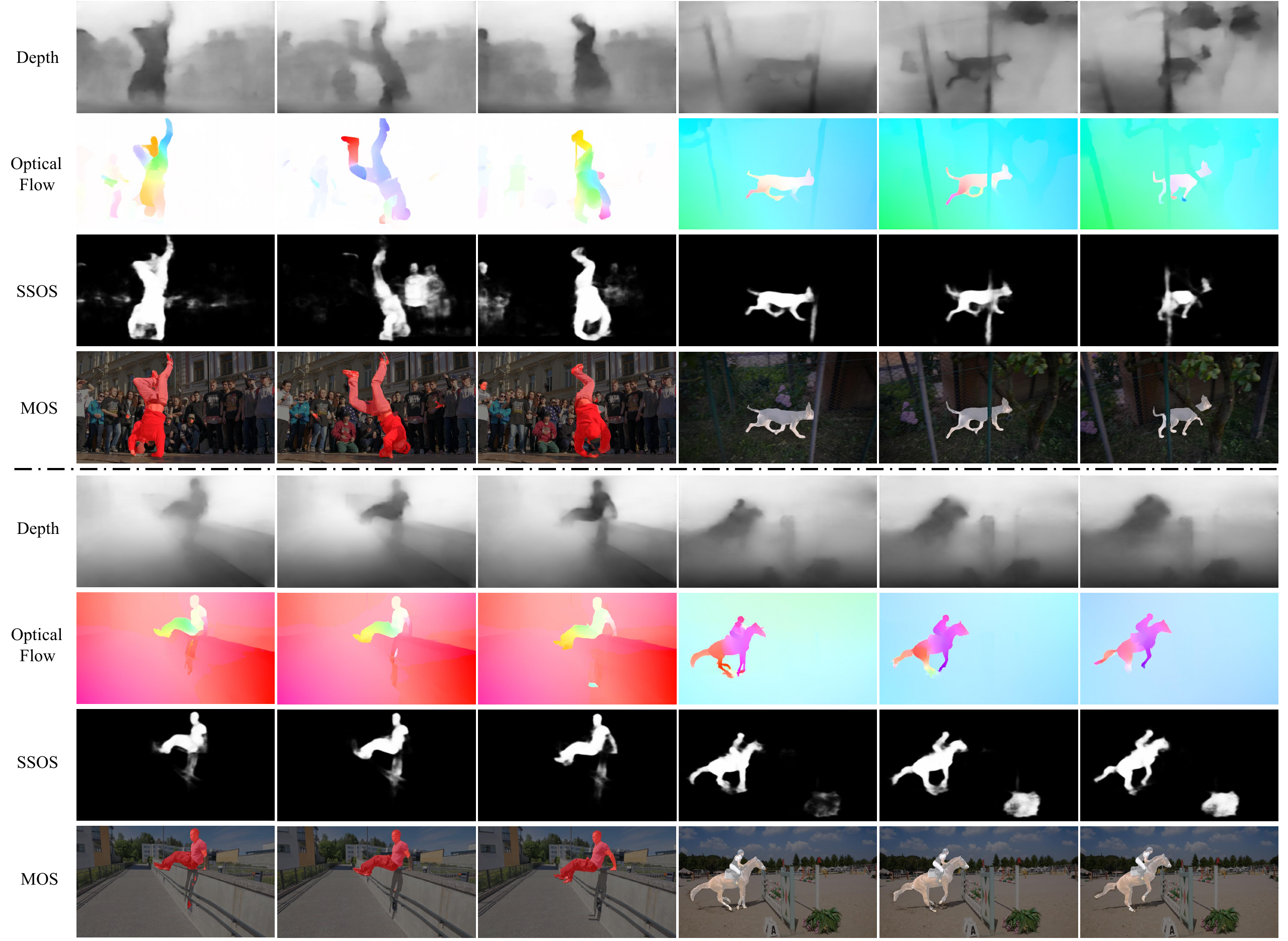}\\ 
    \centering
    \caption{Qualitative results on four sequences $breakdance$, $libby$, $horsejump-high$ and $parkour$ of DAVIS$_{16}$.} 		
    \label{fig:Figure7}
\end{figure*}
\textbf{Performance on FBMS.} In order to make a comprehensive comparison with the previous methods, we also show the results on FBMS, as shown in Tab.~\ref{table:Table3}. Our method still achieves the best performance compared to the others.\\
\textbf{Qualitative Reuslts.}
 Fig.~\ref{fig:Figure7} illustrates visual results of the proposed algorithm on the challenge video sequences $breakdance$, $libby$, $horsejump-high$ and $parkour$ of DAVIS$_{16}$. We can see that each source provides rich location and appearance information. Moreover, depth map can supplement extra contrast information, and the high-quality optical flow can provide clear motion information.  
 Notably, none of these sources can dominate the final prediction, and all of the source characteristics must be integrated to achieve high-precision video object segmentation.
\begin{table}
	\centering
	  \setlength{\abovecaptionskip}{3pt}
	\caption{Ablation study on the validation set of DAVIS$_{16}$.}
	\resizebox{0.45\textwidth}{!}{
		\setlength\tabcolsep{3pt}
		\renewcommand\arraystretch{1.0}
		\begin{tabular}{c|c||cc}
			\hline
			Components &Module &mean $\mathcal{J}$ &mean $\mathcal{F}$\\
			\hline
			\hline 
			\multirow{5}{*}{{Multi-source}}&{RGB} &69.2 &66.7\\
			&{RGB+D} &72.5 &69.0\\
			&{RGB+SOS} &72.9 &69.3\\
	       {feature inputs} &{RGB+OF} &77.3 &76.8\\
	        &{RGB+D+SOS+OF} &80.8 &80.1\\
			\hline
            \hline
			\multirow{2}{*}{{Multi-source  fusion}} & +ISAM & 82.5 & 81.4\\
			& +FPM & 83.4 & 82.3\\
            \hline
		    \hline
			Prediction& +APS  & 83.3 & 82.1\\
			\hline
	\end{tabular}}	
	\label{tab:Table4}
	\vspace{-8mm}
\end{table}
\subsection{Ablation Study}\label{sec:AbStd}
In this section, we detail the contribution of each component to the overall network. Because the optical flow maps always are high quality on the DAVIS$_{16}$. 
We perform an ablation study on it to investigate the effect of multi-source fusion.
Specifically, we first verify the effectiveness of each source for moving object segmentation (MOS). Next, we show the benefits of ISAM and FPM in the multi-source fusion network, respectively. Finally, we evaluate the performance of the automatic predictor selection (APS) on both DAVIS$_{16}$ and Youtube-Objects. 
\begin{figure}[t]
\includegraphics[width=\linewidth]{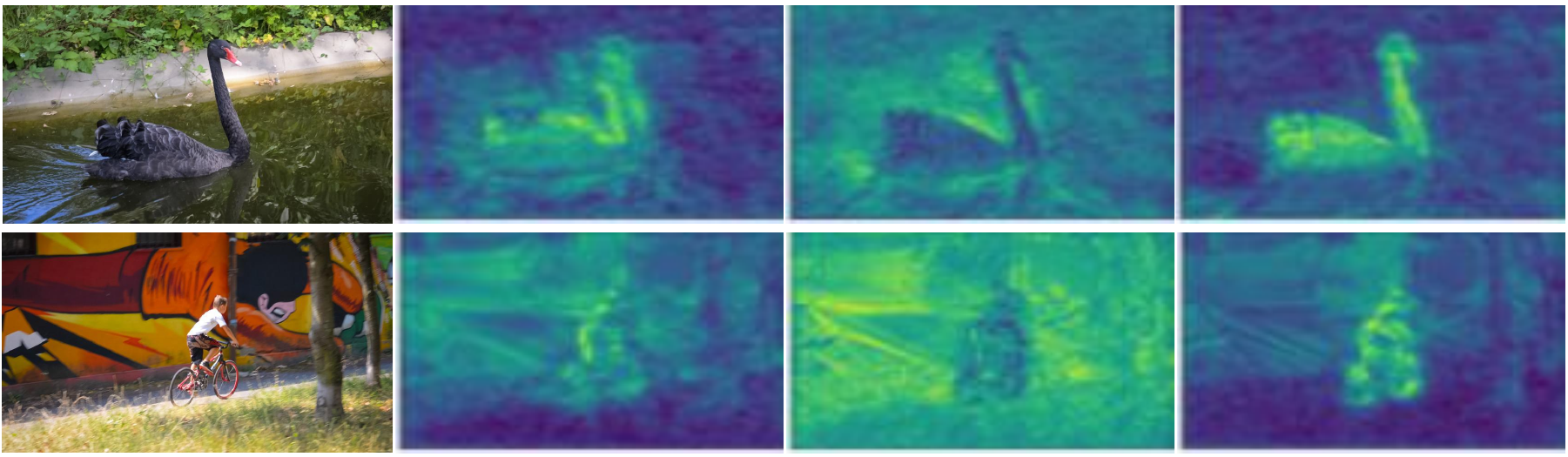}\\
        \centering
        \caption{Visual results  of  ${P}_{comm}^i$, ${P}_{exclu}^i$ and ${P}^i$ for  showing  the  effect of the feature purification module.}
\label{fig:visual_fpm}
\vspace{-5mm}
\end{figure} 

\textbf{Effectiveness of Multi-source Features.} 
We quantitatively show the benefit of each source in Tab.~\ref{tab:Table4}.
We take the FPN with the only RGB feature inputs as the baseline. First, the depth features, static saliency features and optical flow features are added to the baseline network, respectively. It can be seen that the combination of RGB source and other sources has a significant improvement compared to the baseline, with the gain of $16.8\%$ and $20.1\%$ in terms of mean $\mathcal{J}$ and mean $\mathcal{F}$. In the process, multi-source features complement each other without repulsion.
\begin{figure*}
    \includegraphics[width=0.8\textwidth]{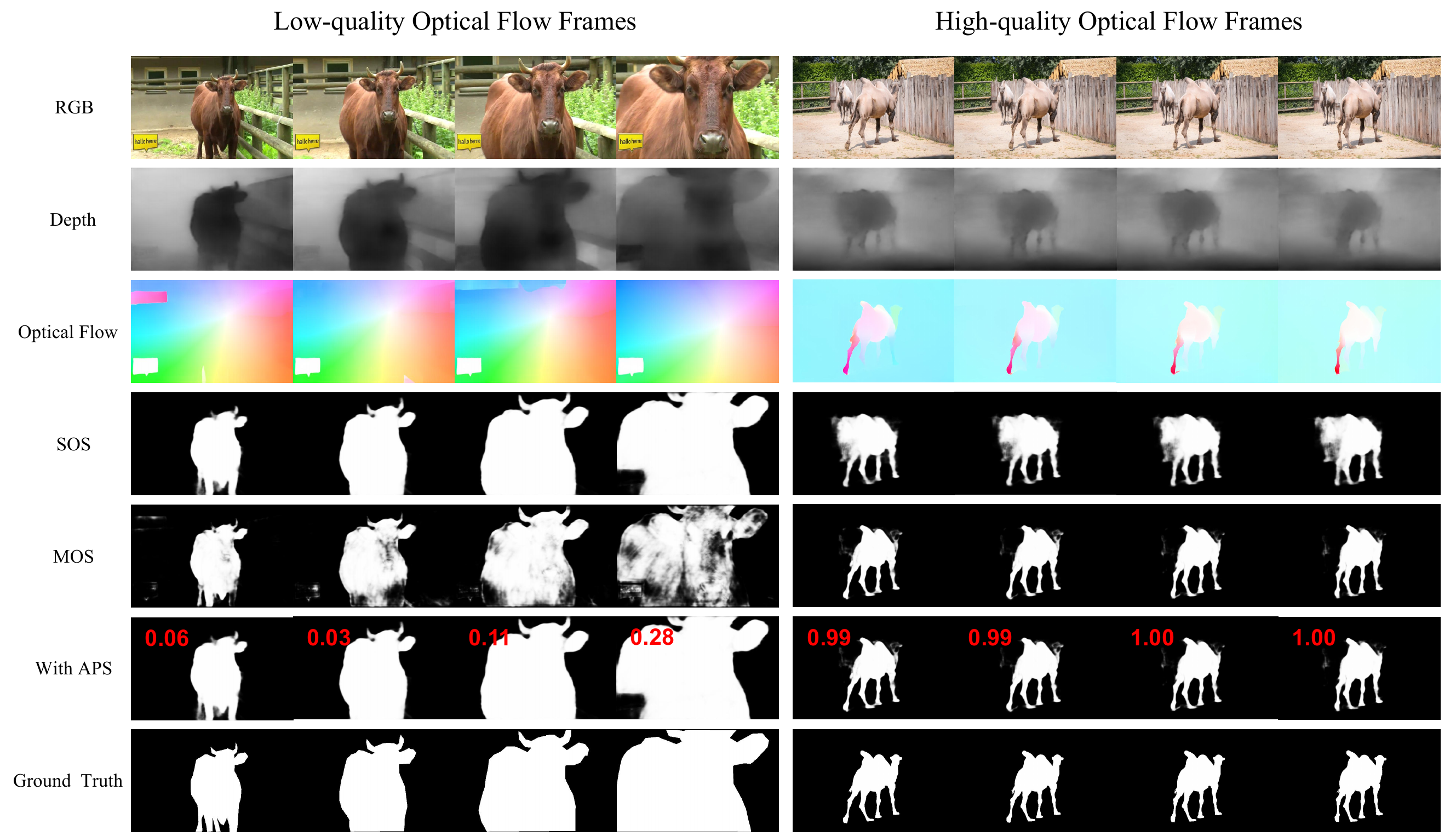}\\ 
    \centering
    \caption{Qualitative results on two example videos $cow$ (low-quality optical flow) and $camel$ (high-quality optical flow), which are from Youtube-Objects and DAVIS$_{16}$, respectively.} 		
    \label{fig:Figure8}
\end{figure*} 

\textbf{Effectiveness of Multi-source Fusion.} We verify the effectiveness of ISAM and FPM in multi-source feature fusion, as shown in Tab.~\ref{tab:Table4}. Compared to directly concatenating multi-source feature maps, the ISAM achieves an improvement of $2.1$\% and $1.6$\% in terms of mean $\mathcal{J}$ and mean $\mathcal{F}$. By further equipping the FPM, it can yield continuous performance improvement of $1.1$\% and $1.1$\% in terms of mean $\mathcal{J}$ and $\mathcal{F}$, respectively (finally obtaining a total of $83.4$ and $82.3$, respectively). To explain the FPM mechanism intuitively, we visualize some features in Fig.~\ref{fig:visual_fpm}. It can be seen that the moving objects in ${P}_{comm}^i$ are disturbed by much background information, whereas ${P}_{exclu}^i$ can perspective the background region well. Their subtraction in the FPM actually builds a kind of information constraint by diverse parameters, thereby guaranteeing that ${P}^i$ can focus on  moving objects.
\begin{table}
	\centering
	\caption{Evaluation of the APS network on both DAVIS$_{16}$ and Youtube-Objects in terms of mean $\mathcal{J}$. }
	\small
	\resizebox{0.38\textwidth}{!}{
		\begin{tabular}{c||c||c}
			\hline
			{Prediction} &	{Youtube-Objects} &{DAVIS$_{16}$}\\
			\hline
			\hline 
			{SOS} &72.7 &67.7\\
			{MOS}  &70.2 &83.4\\
	        {APS}  &74.9 &83.3\\
			\hline
	\end{tabular}}	
	\label{tab:Table5}
	\vspace{-5mm}
\end{table}

\textbf{Effectiveness of Automatic Predictor Selection.} In Tab.~\ref{tab:Table5}, we detail the performance comparison of using APS network on the Youtube-Objects and DAVIS$_{16}$ datasets, respectively. With the help of high-quality optical flow map, the MOS evidently outperforms the SOS on the DAVIS$_{16}$. On the contrary, the low-quality optical flow on the Youtube-Objects encumbers the prediction of the MOS, that is, they are not as good as those of SOS. The designed APS network comprehensively evaluates the degree of interference of optical flow to the MOS. 
%
It can be seen that our three-stage solving strategy significantly outperforms either of the stand-alone SOS or MOS on the Youtube-Objects.
This indicates that the APS is not partial to either of them and has learned the selection mechanism.
%
In addition, the results discriminated by the APS have slightly less mean $\mathcal{J}$ ($0.01$) on the DAVIS$_{16}$ than those generated only by MOS, demonstrating that the discriminator in most times will accept the results assisted with high-quality optical flow maps.
%
To more intuitively show the effectiveness of the APS, we visualize all sources and the results of each stage under optical flow maps of various quality in Fig.~\ref{fig:Figure8}. It can be observed that the APS tends to give low scores to the video sequences with low-quality optical flow and high scores to the video sequences with high-quality optical flow, respectively.

\section{Conclusion}
In this paper, we propose a novel multi-source fusion network to effectively utilize the complementary features from the RGB, depth, static saliency and optical flow for zero-shot video object segmentation. The proposed interoceptive spatial attention module can adaptively enhance the feature representation of each source in spatial position. With the help of feature purification module, it can further filter the incompatible features among sources to refine the multi-source fused features. In addition, to get rid of the inevitable interference caused by low-quality optical flow, we design a novel predictor selection network, which automatically chooses the results from static saliency predictor and moving object predictor.
%
The automatic predictor selection network is simple yet effective, therefore, it can provide an important reference for other optical flow based methods. Extensive experiments on three ZVOS datasets indicate that the proposed method performs favorably against the current state-of-the-art methods.


\begin{acks}
This work was supported in part by the National Natural Science Foundation of China \#61876202, \#61725202, \#61751212 and \#61829102,
the Dalian Science and Technology Innovation Foundation \#2019J12GX039, 
and the Fundamental Research Funds for the Central Universities \#DUT20ZD212. 
\end{acks}




\bibliographystyle{ACM-Reference-Format}
\bibliography{acmart.bib}


\begin{thebibliography}{59}


\ifx \showCODEN    \undefined \def \showCODEN     #1{\unskip}     \fi
\ifx \showDOI      \undefined \def \showDOI       #1{#1}\fi
\ifx \showISBNx    \undefined \def \showISBNx     #1{\unskip}     \fi
\ifx \showISBNxiii \undefined \def \showISBNxiii  #1{\unskip}     \fi
\ifx \showISSN     \undefined \def \showISSN      #1{\unskip}     \fi
\ifx \showLCCN     \undefined \def \showLCCN      #1{\unskip}     \fi
\ifx \shownote     \undefined \def \shownote      #1{#1}          \fi
\ifx \showarticletitle \undefined \def \showarticletitle #1{#1}   \fi
\ifx \showURL      \undefined \def \showURL       {\relax}        \fi
\providecommand\bibfield[2]{#2}
\providecommand\bibinfo[2]{#2}
\providecommand\natexlab[1]{#1}
\providecommand\showeprint[2][]{arXiv:#2}

\bibitem[\protect\citeauthoryear{An, Zhao, and Hou}{An et~al\mbox{.}}{2016}]%
        {rgbd-tracking3}
\bibfield{author}{\bibinfo{person}{Ning An}, \bibinfo{person}{Xiao-Guang Zhao},
  {and} \bibinfo{person}{Zeng-Guang Hou}.} \bibinfo{year}{2016}\natexlab{}.
\newblock \showarticletitle{Online RGB-D tracking via
  detection-learning-segmentation}. In \bibinfo{booktitle}{\emph{ICPR}}.
  \bibinfo{pages}{1231--1236}.
\newblock


\bibitem[\protect\citeauthoryear{Cheng, Tsai, Wang, and Yang}{Cheng
  et~al\mbox{.}}{2017}]%
        {SFL}
\bibfield{author}{\bibinfo{person}{Jingchun Cheng}, \bibinfo{person}{Yi-Hsuan
  Tsai}, \bibinfo{person}{Shengjin Wang}, {and} \bibinfo{person}{Ming-Hsuan
  Yang}.} \bibinfo{year}{2017}\natexlab{}.
\newblock \showarticletitle{Segflow: Joint learning for video object
  segmentation and optical flow}. In \bibinfo{booktitle}{\emph{ICCV}}.
  \bibinfo{pages}{686--695}.
\newblock


\bibitem[\protect\citeauthoryear{Cheng, Fu, Wei, Xiao, and Cao}{Cheng
  et~al\mbox{.}}{2014}]%
        {RGBD135}
\bibfield{author}{\bibinfo{person}{Yupeng Cheng}, \bibinfo{person}{Huazhu Fu},
  \bibinfo{person}{Xingxing Wei}, \bibinfo{person}{Jiangjian Xiao}, {and}
  \bibinfo{person}{Xiaochun Cao}.} \bibinfo{year}{2014}\natexlab{}.
\newblock \showarticletitle{Depth enhanced saliency detection method}. In
  \bibinfo{booktitle}{\emph{ICIMCS}}. \bibinfo{pages}{23}.
\newblock


\bibitem[\protect\citeauthoryear{De~Boer, Kroese, Mannor, and
  Rubinstein}{De~Boer et~al\mbox{.}}{2005}]%
        {BCE}
\bibfield{author}{\bibinfo{person}{Pieter-Tjerk De~Boer},
  \bibinfo{person}{Dirk~P Kroese}, \bibinfo{person}{Shie Mannor}, {and}
  \bibinfo{person}{Reuven~Y Rubinstein}.} \bibinfo{year}{2005}\natexlab{}.
\newblock \showarticletitle{A tutorial on the cross-entropy method}.
\newblock \bibinfo{journal}{\emph{Annals of operations research}}
  \bibinfo{volume}{134}, \bibinfo{number}{1} (\bibinfo{year}{2005}),
  \bibinfo{pages}{19--67}.
\newblock


\bibitem[\protect\citeauthoryear{Deng, Dong, Socher, Li, Li, and Fei-Fei}{Deng
  et~al\mbox{.}}{2009}]%
        {ImageNet}
\bibfield{author}{\bibinfo{person}{Jia Deng}, \bibinfo{person}{Wei Dong},
  \bibinfo{person}{Richard Socher}, \bibinfo{person}{Li-Jia Li},
  \bibinfo{person}{Kai Li}, {and} \bibinfo{person}{Li Fei-Fei}.}
  \bibinfo{year}{2009}\natexlab{}.
\newblock \showarticletitle{Imagenet: A large-scale hierarchical image
  database}. In \bibinfo{booktitle}{\emph{CVPR}}. \bibinfo{pages}{248--255}.
\newblock


\bibitem[\protect\citeauthoryear{Faisal, Akhter, Ali, and Hartley}{Faisal
  et~al\mbox{.}}{2019}]%
        {EPO}
\bibfield{author}{\bibinfo{person}{Muhammad Faisal}, \bibinfo{person}{Ijaz
  Akhter}, \bibinfo{person}{Mohsen Ali}, {and} \bibinfo{person}{Richard
  Hartley}.} \bibinfo{year}{2019}\natexlab{}.
\newblock \showarticletitle{Exploiting geometric constraints on dense
  trajectories for motion saliency}.
\newblock \bibinfo{journal}{\emph{arXiv preprint arXiv:1909.13258}}
  (\bibinfo{year}{2019}).
\newblock


\bibitem[\protect\citeauthoryear{Fan, Lin, Zhao, Liu, Zhang, Hou, Zhu, and
  Cheng}{Fan et~al\mbox{.}}{2019}]%
        {SIP}
\bibfield{author}{\bibinfo{person}{Deng-Ping Fan}, \bibinfo{person}{Zheng Lin},
  \bibinfo{person}{Jia-Xing Zhao}, \bibinfo{person}{Yun Liu},
  \bibinfo{person}{Zhao Zhang}, \bibinfo{person}{Qibin Hou},
  \bibinfo{person}{Menglong Zhu}, {and} \bibinfo{person}{Ming-Ming Cheng}.}
  \bibinfo{year}{2019}\natexlab{}.
\newblock \showarticletitle{Rethinking RGB-D salient object detection: Models,
  datasets, and large-scale benchmarks}.
\newblock \bibinfo{journal}{\emph{arXiv preprint arXiv:1907.06781}}
  (\bibinfo{year}{2019}).
\newblock


\bibitem[\protect\citeauthoryear{He, Zhang, Ren, and Sun}{He
  et~al\mbox{.}}{2015}]%
        {PRelu}
\bibfield{author}{\bibinfo{person}{Kaiming He}, \bibinfo{person}{Xiangyu
  Zhang}, \bibinfo{person}{Shaoqing Ren}, {and} \bibinfo{person}{Jian Sun}.}
  \bibinfo{year}{2015}\natexlab{}.
\newblock \showarticletitle{Delving deep into rectifiers: Surpassing
  human-level performance on imagenet classification}. In
  \bibinfo{booktitle}{\emph{ICCV}}. \bibinfo{pages}{1026--1034}.
\newblock


\bibitem[\protect\citeauthoryear{He, Zhang, Ren, and Sun}{He
  et~al\mbox{.}}{2016}]%
        {Resnet}
\bibfield{author}{\bibinfo{person}{Kaiming He}, \bibinfo{person}{Xiangyu
  Zhang}, \bibinfo{person}{Shaoqing Ren}, {and} \bibinfo{person}{Jian Sun}.}
  \bibinfo{year}{2016}\natexlab{}.
\newblock \showarticletitle{Deep residual learning for image recognition}. In
  \bibinfo{booktitle}{\emph{CVPR}}. \bibinfo{pages}{770--778}.
\newblock


\bibitem[\protect\citeauthoryear{Hou, Cheng, Hu, Borji, Tu, and Torr}{Hou
  et~al\mbox{.}}{2017}]%
        {DSS}
\bibfield{author}{\bibinfo{person}{Qibin Hou}, \bibinfo{person}{Ming-Ming
  Cheng}, \bibinfo{person}{Xiaowei Hu}, \bibinfo{person}{Ali Borji},
  \bibinfo{person}{Zhuowen Tu}, {and} \bibinfo{person}{Philip~HS Torr}.}
  \bibinfo{year}{2017}\natexlab{}.
\newblock \showarticletitle{Deeply supervised salient object detection with
  short connections}. In \bibinfo{booktitle}{\emph{CVPR}}.
  \bibinfo{pages}{3203--3212}.
\newblock


\bibitem[\protect\citeauthoryear{Hui, Tang, and Change~Loy}{Hui
  et~al\mbox{.}}{2018}]%
        {liteflownet}
\bibfield{author}{\bibinfo{person}{Tak-Wai Hui}, \bibinfo{person}{Xiaoou Tang},
  {and} \bibinfo{person}{Chen Change~Loy}.} \bibinfo{year}{2018}\natexlab{}.
\newblock \showarticletitle{Liteflownet: A lightweight convolutional neural
  network for optical flow estimation}. In \bibinfo{booktitle}{\emph{CVPR}}.
  \bibinfo{pages}{8981--8989}.
\newblock


\bibitem[\protect\citeauthoryear{Jain, Xiong, and Grauman}{Jain
  et~al\mbox{.}}{2017}]%
        {FSEG}
\bibfield{author}{\bibinfo{person}{Suyog~Dutt Jain}, \bibinfo{person}{Bo
  Xiong}, {and} \bibinfo{person}{Kristen Grauman}.}
  \bibinfo{year}{2017}\natexlab{}.
\newblock \showarticletitle{Fusionseg: Learning to combine motion and
  appearance for fully automatic segmentation of generic objects in videos}. In
  \bibinfo{booktitle}{\emph{CVPR}}. \bibinfo{pages}{2117--2126}.
\newblock


\bibitem[\protect\citeauthoryear{Ji, Jia, Lu, and Ruan}{Ji
  et~al\mbox{.}}{2021}]%
        {jiyuan_mm}
\bibfield{author}{\bibinfo{person}{Yuan Ji}, \bibinfo{person}{Xu Jia},
  \bibinfo{person}{Huchuan Lu}, {and} \bibinfo{person}{Xiang Ruan}.}
  \bibinfo{year}{2021}\natexlab{}.
\newblock \showarticletitle{Weakly-Supervised Temporal Action Localization via
  Cross-Stream Collaborative Learning}. In \bibinfo{booktitle}{\emph{ACMMM}}.
\newblock


\bibitem[\protect\citeauthoryear{Ju, Ge, Geng, Ren, and Wu}{Ju
  et~al\mbox{.}}{2014}]%
        {NJU2000}
\bibfield{author}{\bibinfo{person}{Ran Ju}, \bibinfo{person}{Ling Ge},
  \bibinfo{person}{Wenjing Geng}, \bibinfo{person}{Tongwei Ren}, {and}
  \bibinfo{person}{Gangshan Wu}.} \bibinfo{year}{2014}\natexlab{}.
\newblock \showarticletitle{Depth saliency based on anisotropic center-surround
  difference}. In \bibinfo{booktitle}{\emph{ICIP}}.
  \bibinfo{pages}{1115--1119}.
\newblock


\bibitem[\protect\citeauthoryear{Jun~Koh and Kim}{Jun~Koh and Kim}{2017}]%
        {ARP}
\bibfield{author}{\bibinfo{person}{Yeong Jun~Koh} {and}
  \bibinfo{person}{Chang-Su Kim}.} \bibinfo{year}{2017}\natexlab{}.
\newblock \showarticletitle{Primary object segmentation in videos based on
  region augmentation and reduction}. In \bibinfo{booktitle}{\emph{CVPR}}.
  \bibinfo{pages}{3442--3450}.
\newblock


\bibitem[\protect\citeauthoryear{Li, Seybold, Vorobyov, Lei, and Jay~Kuo}{Li
  et~al\mbox{.}}{2018}]%
        {UVOS-Bilateral}
\bibfield{author}{\bibinfo{person}{Siyang Li}, \bibinfo{person}{Bryan Seybold},
  \bibinfo{person}{Alexey Vorobyov}, \bibinfo{person}{Xuejing Lei}, {and}
  \bibinfo{person}{C-C Jay~Kuo}.} \bibinfo{year}{2018}\natexlab{}.
\newblock \showarticletitle{Unsupervised video object segmentation with
  motion-based bilateral networks}. In \bibinfo{booktitle}{\emph{ECCV}}.
  \bibinfo{pages}{207--223}.
\newblock


\bibitem[\protect\citeauthoryear{Lin, Doll{\'a}r, Girshick, He, Hariharan, and
  Belongie}{Lin et~al\mbox{.}}{2017}]%
        {FPN}
\bibfield{author}{\bibinfo{person}{Tsung-Yi Lin}, \bibinfo{person}{Piotr
  Doll{\'a}r}, \bibinfo{person}{Ross Girshick}, \bibinfo{person}{Kaiming He},
  \bibinfo{person}{Bharath Hariharan}, {and} \bibinfo{person}{Serge Belongie}.}
  \bibinfo{year}{2017}\natexlab{}.
\newblock \showarticletitle{Feature pyramid networks for object detection}. In
  \bibinfo{booktitle}{\emph{CVPR}}. \bibinfo{pages}{2117--2125}.
\newblock


\bibitem[\protect\citeauthoryear{Liu, Wu, Wang, and Qian}{Liu
  et~al\mbox{.}}{2018}]%
        {SS-RGBD-SS}
\bibfield{author}{\bibinfo{person}{Hong Liu}, \bibinfo{person}{Wenshan Wu},
  \bibinfo{person}{Xiangdong Wang}, {and} \bibinfo{person}{Yueliang Qian}.}
  \bibinfo{year}{2018}\natexlab{}.
\newblock \showarticletitle{RGB-D joint modelling with scene geometric
  information for indoor semantic segmentation}.
\newblock \bibinfo{journal}{\emph{Multimedia Tools and Applications}}
  \bibinfo{volume}{77}, \bibinfo{number}{17} (\bibinfo{year}{2018}),
  \bibinfo{pages}{22475--22488}.
\newblock


\bibitem[\protect\citeauthoryear{Liu, Rabinovich, and Berg}{Liu
  et~al\mbox{.}}{2015}]%
        {poly}
\bibfield{author}{\bibinfo{person}{Wei Liu}, \bibinfo{person}{Andrew
  Rabinovich}, {and} \bibinfo{person}{Alexander~C Berg}.}
  \bibinfo{year}{2015}\natexlab{}.
\newblock \showarticletitle{Parsenet: Looking wider to see better}.
\newblock \bibinfo{journal}{\emph{arXiv preprint arXiv:1506.04579}}
  (\bibinfo{year}{2015}).
\newblock


\bibitem[\protect\citeauthoryear{Lu, Wang, Ma, Shen, Shao, and Porikli}{Lu
  et~al\mbox{.}}{2019}]%
        {COSNet}
\bibfield{author}{\bibinfo{person}{Xiankai Lu}, \bibinfo{person}{Wenguan Wang},
  \bibinfo{person}{Chao Ma}, \bibinfo{person}{Jianbing Shen},
  \bibinfo{person}{Ling Shao}, {and} \bibinfo{person}{Fatih Porikli}.}
  \bibinfo{year}{2019}\natexlab{}.
\newblock \showarticletitle{See more, know more: Unsupervised video object
  segmentation with co-attention siamese networks}. In
  \bibinfo{booktitle}{\emph{CVPR}}. \bibinfo{pages}{3623--3632}.
\newblock


\bibitem[\protect\citeauthoryear{Lukezic, Kart, Kapyla, Durmush, Kamarainen,
  Matas, and Kristan}{Lukezic et~al\mbox{.}}{2019}]%
        {rgbd-tracking1}
\bibfield{author}{\bibinfo{person}{Alan Lukezic}, \bibinfo{person}{Ugur Kart},
  \bibinfo{person}{Jani Kapyla}, \bibinfo{person}{Ahmed Durmush},
  \bibinfo{person}{Joni-Kristian Kamarainen}, \bibinfo{person}{Jiri Matas},
  {and} \bibinfo{person}{Matej Kristan}.} \bibinfo{year}{2019}\natexlab{}.
\newblock \showarticletitle{CDTB: A color and depth visual object tracking
  dataset and benchmark}. In \bibinfo{booktitle}{\emph{ICCV}}.
  \bibinfo{pages}{10013--10022}.
\newblock


\bibitem[\protect\citeauthoryear{Niu, Geng, Li, and Liu}{Niu
  et~al\mbox{.}}{2012}]%
        {STERE}
\bibfield{author}{\bibinfo{person}{Yuzhen Niu}, \bibinfo{person}{Yujie Geng},
  \bibinfo{person}{Xueqing Li}, {and} \bibinfo{person}{Feng Liu}.}
  \bibinfo{year}{2012}\natexlab{}.
\newblock \showarticletitle{Leveraging stereopsis for saliency analysis}. In
  \bibinfo{booktitle}{\emph{CVPR}}. \bibinfo{pages}{454--461}.
\newblock


\bibitem[\protect\citeauthoryear{Ocal and Mustafa}{Ocal and Mustafa}{2020}]%
        {depth5}
\bibfield{author}{\bibinfo{person}{Mertalp Ocal} {and} \bibinfo{person}{Armin
  Mustafa}.} \bibinfo{year}{2020}\natexlab{}.
\newblock \showarticletitle{RealMonoDepth: Self-Supervised Monocular Depth
  Estimation for General Scenes}.
\newblock \bibinfo{journal}{\emph{arXiv preprint arXiv:2004.06267}}
  (\bibinfo{year}{2020}).
\newblock


\bibitem[\protect\citeauthoryear{Ochs, Malik, and Brox}{Ochs
  et~al\mbox{.}}{2013}]%
        {FBMS}
\bibfield{author}{\bibinfo{person}{Peter Ochs}, \bibinfo{person}{Jitendra
  Malik}, {and} \bibinfo{person}{Thomas Brox}.}
  \bibinfo{year}{2013}\natexlab{}.
\newblock \showarticletitle{Segmentation of moving objects by long term video
  analysis}.
\newblock \bibinfo{journal}{\emph{IEEE TPAMI}} \bibinfo{volume}{36},
  \bibinfo{number}{6} (\bibinfo{year}{2013}), \bibinfo{pages}{1187--1200}.
\newblock


\bibitem[\protect\citeauthoryear{Pang, Zhang, Zhao, and Lu}{Pang
  et~al\mbox{.}}{2020a}]%
        {HDFNet}
\bibfield{author}{\bibinfo{person}{Youwei Pang}, \bibinfo{person}{Lihe Zhang},
  \bibinfo{person}{Xiaoqi Zhao}, {and} \bibinfo{person}{Huchuan Lu}.}
  \bibinfo{year}{2020}\natexlab{a}.
\newblock \showarticletitle{Hierarchical dynamic filtering network for RGB-D
  salient object detection}. In \bibinfo{booktitle}{\emph{ECCV}}.
  \bibinfo{pages}{235--252}.
\newblock


\bibitem[\protect\citeauthoryear{Pang, Zhao, Zhang, and Lu}{Pang
  et~al\mbox{.}}{2020b}]%
        {MINet}
\bibfield{author}{\bibinfo{person}{Youwei Pang}, \bibinfo{person}{Xiaoqi Zhao},
  \bibinfo{person}{Lihe Zhang}, {and} \bibinfo{person}{Huchuan Lu}.}
  \bibinfo{year}{2020}\natexlab{b}.
\newblock \showarticletitle{Multi-Scale Interactive Network for Salient Object
  Detection}. In \bibinfo{booktitle}{\emph{CVPR}}. \bibinfo{pages}{9413--9422}.
\newblock


\bibitem[\protect\citeauthoryear{Papazoglou and Ferrari}{Papazoglou and
  Ferrari}{2013}]%
        {FST}
\bibfield{author}{\bibinfo{person}{Anestis Papazoglou} {and}
  \bibinfo{person}{Vittorio Ferrari}.} \bibinfo{year}{2013}\natexlab{}.
\newblock \showarticletitle{Fast object segmentation in unconstrained video}.
  In \bibinfo{booktitle}{\emph{ICCV}}. \bibinfo{pages}{1777--1784}.
\newblock


\bibitem[\protect\citeauthoryear{Peng, Li, Xiong, Hu, and Ji}{Peng
  et~al\mbox{.}}{2014}]%
        {early_fusion_1}
\bibfield{author}{\bibinfo{person}{Houwen Peng}, \bibinfo{person}{Bing Li},
  \bibinfo{person}{Weihua Xiong}, \bibinfo{person}{Weiming Hu}, {and}
  \bibinfo{person}{Rongrong Ji}.} \bibinfo{year}{2014}\natexlab{}.
\newblock \showarticletitle{RGBD salient object detection: A benchmark and
  algorithms}. In \bibinfo{booktitle}{\emph{ECCV}}. \bibinfo{pages}{92--109}.
\newblock


\bibitem[\protect\citeauthoryear{Perazzi, Pont-Tuset, McWilliams, Van~Gool,
  Gross, and Sorkine-Hornung}{Perazzi et~al\mbox{.}}{2016}]%
        {davis16}
\bibfield{author}{\bibinfo{person}{Federico Perazzi}, \bibinfo{person}{Jordi
  Pont-Tuset}, \bibinfo{person}{Brian McWilliams}, \bibinfo{person}{Luc
  Van~Gool}, \bibinfo{person}{Markus Gross}, {and} \bibinfo{person}{Alexander
  Sorkine-Hornung}.} \bibinfo{year}{2016}\natexlab{}.
\newblock \showarticletitle{A benchmark dataset and evaluation methodology for
  video object segmentation}. In \bibinfo{booktitle}{\emph{CVPR}}.
  \bibinfo{pages}{724--732}.
\newblock


\bibitem[\protect\citeauthoryear{Piao, Ji, Li, Zhang, and Lu}{Piao
  et~al\mbox{.}}{2019}]%
        {DMRA}
\bibfield{author}{\bibinfo{person}{Yongri Piao}, \bibinfo{person}{Wei Ji},
  \bibinfo{person}{Jingjing Li}, \bibinfo{person}{Miao Zhang}, {and}
  \bibinfo{person}{Huchuan Lu}.} \bibinfo{year}{2019}\natexlab{}.
\newblock \showarticletitle{Depth-Induced Multi-Scale Recurrent Attention
  Network for Saliency Detection}. In \bibinfo{booktitle}{\emph{ICCV}}.
  \bibinfo{pages}{7254--7263}.
\newblock


\bibitem[\protect\citeauthoryear{Pillai, Ambru{\c{s}}, and Gaidon}{Pillai
  et~al\mbox{.}}{2019}]%
        {depth3}
\bibfield{author}{\bibinfo{person}{Sudeep Pillai}, \bibinfo{person}{Rare{\c{s}}
  Ambru{\c{s}}}, {and} \bibinfo{person}{Adrien Gaidon}.}
  \bibinfo{year}{2019}\natexlab{}.
\newblock \showarticletitle{Superdepth: Self-supervised, super-resolved
  monocular depth estimation}. In \bibinfo{booktitle}{\emph{ICRA}}.
  \bibinfo{pages}{9250--9256}.
\newblock


\bibitem[\protect\citeauthoryear{Prest, Leistner, Civera, Schmid, and
  Ferrari}{Prest et~al\mbox{.}}{2012}]%
        {youtube-objects}
\bibfield{author}{\bibinfo{person}{Alessandro Prest},
  \bibinfo{person}{Christian Leistner}, \bibinfo{person}{Javier Civera},
  \bibinfo{person}{Cordelia Schmid}, {and} \bibinfo{person}{Vittorio Ferrari}.}
  \bibinfo{year}{2012}\natexlab{}.
\newblock \showarticletitle{Learning object class detectors from weakly
  annotated video}. In \bibinfo{booktitle}{\emph{CVPR}}.
  \bibinfo{pages}{3282--3289}.
\newblock


\bibitem[\protect\citeauthoryear{Qin, Zhang, Huang, Gao, Dehghan, and
  Jagersand}{Qin et~al\mbox{.}}{2019}]%
        {BASNet}
\bibfield{author}{\bibinfo{person}{Xuebin Qin}, \bibinfo{person}{Zichen Zhang},
  \bibinfo{person}{Chenyang Huang}, \bibinfo{person}{Chao Gao},
  \bibinfo{person}{Masood Dehghan}, {and} \bibinfo{person}{Martin Jagersand}.}
  \bibinfo{year}{2019}\natexlab{}.
\newblock \showarticletitle{BASNet: Boundary-Aware Salient Object Detection}.
  In \bibinfo{booktitle}{\emph{CVPR}}. \bibinfo{pages}{7479--7489}.
\newblock


\bibitem[\protect\citeauthoryear{Ranftl, Lasinger, Hafner, Schindler, and
  Koltun}{Ranftl et~al\mbox{.}}{2020}]%
        {depth4}
\bibfield{author}{\bibinfo{person}{Ren{\'e} Ranftl}, \bibinfo{person}{Katrin
  Lasinger}, \bibinfo{person}{David Hafner}, \bibinfo{person}{Konrad
  Schindler}, {and} \bibinfo{person}{Vladlen Koltun}.}
  \bibinfo{year}{2020}\natexlab{}.
\newblock \showarticletitle{Towards robust monocular depth estimation: Mixing
  datasets for zero-shot cross-dataset transfer}.
\newblock \bibinfo{journal}{\emph{IEEE TPAMI}} (\bibinfo{year}{2020}).
\newblock


\bibitem[\protect\citeauthoryear{Ranjan and Black}{Ranjan and Black}{2017}]%
        {SFN}
\bibfield{author}{\bibinfo{person}{Anurag Ranjan} {and}
  \bibinfo{person}{Michael~J Black}.} \bibinfo{year}{2017}\natexlab{}.
\newblock \showarticletitle{Optical flow estimation using a spatial pyramid
  network}. In \bibinfo{booktitle}{\emph{CVPR}}. \bibinfo{pages}{4161--4170}.
\newblock


\bibitem[\protect\citeauthoryear{Rasoulidanesh, Yadav, Herath, Vaghei, and
  Payandeh}{Rasoulidanesh et~al\mbox{.}}{2019}]%
        {rgbd-tracking2}
\bibfield{author}{\bibinfo{person}{Maryamsadat Rasoulidanesh},
  \bibinfo{person}{Srishti Yadav}, \bibinfo{person}{Sachini Herath},
  \bibinfo{person}{Yasaman Vaghei}, {and} \bibinfo{person}{Shahram Payandeh}.}
  \bibinfo{year}{2019}\natexlab{}.
\newblock \showarticletitle{Deep Attention Models for Human Tracking Using
  RGBD}.
\newblock \bibinfo{journal}{\emph{Sensors}} \bibinfo{volume}{19},
  \bibinfo{number}{4} (\bibinfo{year}{2019}), \bibinfo{pages}{750}.
\newblock


\bibitem[\protect\citeauthoryear{Siam, Jiang, Lu, Petrich, Gamal, Elhoseiny,
  and Jagersand}{Siam et~al\mbox{.}}{2019}]%
        {MotAdapt}
\bibfield{author}{\bibinfo{person}{Mennatullah Siam}, \bibinfo{person}{Chen
  Jiang}, \bibinfo{person}{Steven Lu}, \bibinfo{person}{Laura Petrich},
  \bibinfo{person}{Mahmoud Gamal}, \bibinfo{person}{Mohamed Elhoseiny}, {and}
  \bibinfo{person}{Martin Jagersand}.} \bibinfo{year}{2019}\natexlab{}.
\newblock \showarticletitle{Video object segmentation using teacher-student
  adaptation in a human robot interaction (hri) setting}. In
  \bibinfo{booktitle}{\emph{ICRA}}. \bibinfo{pages}{50--56}.
\newblock


\bibitem[\protect\citeauthoryear{Song, Wang, Zhao, Shen, and Lam}{Song
  et~al\mbox{.}}{2018}]%
        {PDB}
\bibfield{author}{\bibinfo{person}{Hongmei Song}, \bibinfo{person}{Wenguan
  Wang}, \bibinfo{person}{Sanyuan Zhao}, \bibinfo{person}{Jianbing Shen}, {and}
  \bibinfo{person}{Kin-Man Lam}.} \bibinfo{year}{2018}\natexlab{}.
\newblock \showarticletitle{Pyramid dilated deeper convlstm for video salient
  object detection}. In \bibinfo{booktitle}{\emph{ECCV}}.
  \bibinfo{pages}{715--731}.
\newblock


\bibitem[\protect\citeauthoryear{Sun, Yang, Liu, and Kautz}{Sun
  et~al\mbox{.}}{2018}]%
        {PWC}
\bibfield{author}{\bibinfo{person}{D Sun}, \bibinfo{person}{X Yang},
  \bibinfo{person}{MY Liu}, {and} \bibinfo{person}{J Kautz}.}
  \bibinfo{year}{2018}\natexlab{}.
\newblock \showarticletitle{PWC-Net: CNNs for Optical Flow Using Pyramid,
  Warping, and Cost Volume}. In \bibinfo{booktitle}{\emph{CVPR}}.
  \bibinfo{pages}{8934–8943}.
\newblock


\bibitem[\protect\citeauthoryear{Teed and Deng}{Teed and Deng}{2020}]%
        {RAFT}
\bibfield{author}{\bibinfo{person}{Zachary Teed} {and} \bibinfo{person}{Jia
  Deng}.} \bibinfo{year}{2020}\natexlab{}.
\newblock \showarticletitle{Raft: Recurrent all-pairs field transforms for
  optical flow}. In \bibinfo{booktitle}{\emph{ECCV}}.
  \bibinfo{pages}{402--419}.
\newblock


\bibitem[\protect\citeauthoryear{Tokmakov, Alahari, and Schmid}{Tokmakov
  et~al\mbox{.}}{2017a}]%
        {MP}
\bibfield{author}{\bibinfo{person}{Pavel Tokmakov}, \bibinfo{person}{Karteek
  Alahari}, {and} \bibinfo{person}{Cordelia Schmid}.}
  \bibinfo{year}{2017}\natexlab{a}.
\newblock \showarticletitle{Learning motion patterns in videos}. In
  \bibinfo{booktitle}{\emph{CVPR}}. \bibinfo{pages}{3386--3394}.
\newblock


\bibitem[\protect\citeauthoryear{Tokmakov, Alahari, and Schmid}{Tokmakov
  et~al\mbox{.}}{2017b}]%
        {LVO}
\bibfield{author}{\bibinfo{person}{Pavel Tokmakov}, \bibinfo{person}{Karteek
  Alahari}, {and} \bibinfo{person}{Cordelia Schmid}.}
  \bibinfo{year}{2017}\natexlab{b}.
\newblock \showarticletitle{Learning video object segmentation with visual
  memory}. In \bibinfo{booktitle}{\emph{ICCV}}. \bibinfo{pages}{4481--4490}.
\newblock


\bibitem[\protect\citeauthoryear{Tsai, Zhong, and Yang}{Tsai
  et~al\mbox{.}}{2016}]%
        {COSEG}
\bibfield{author}{\bibinfo{person}{Yi-Hsuan Tsai}, \bibinfo{person}{Guangyu
  Zhong}, {and} \bibinfo{person}{Ming-Hsuan Yang}.}
  \bibinfo{year}{2016}\natexlab{}.
\newblock \showarticletitle{Semantic co-segmentation in videos}. In
  \bibinfo{booktitle}{\emph{ECCV}}. \bibinfo{pages}{760--775}.
\newblock


\bibitem[\protect\citeauthoryear{Wang, Zhang, Wang, Lu, Yang, Ruan, and
  Borji}{Wang et~al\mbox{.}}{2018}]%
        {DGRL}
\bibfield{author}{\bibinfo{person}{Tiantian Wang}, \bibinfo{person}{Lihe
  Zhang}, \bibinfo{person}{Shuo Wang}, \bibinfo{person}{Huchuan Lu},
  \bibinfo{person}{Gang Yang}, \bibinfo{person}{Xiang Ruan}, {and}
  \bibinfo{person}{Ali Borji}.} \bibinfo{year}{2018}\natexlab{}.
\newblock \showarticletitle{Detect globally, refine locally: A novel approach
  to saliency detection}. In \bibinfo{booktitle}{\emph{CVPR}}.
  \bibinfo{pages}{3127--3135}.
\newblock


\bibitem[\protect\citeauthoryear{Wang, Lu, Shen, Crandall, and Shao}{Wang
  et~al\mbox{.}}{2019a}]%
        {AGNN}
\bibfield{author}{\bibinfo{person}{Wenguan Wang}, \bibinfo{person}{Xiankai Lu},
  \bibinfo{person}{Jianbing Shen}, \bibinfo{person}{David~J Crandall}, {and}
  \bibinfo{person}{Ling Shao}.} \bibinfo{year}{2019}\natexlab{a}.
\newblock \showarticletitle{Zero-shot video object segmentation via attentive
  graph neural networks}. In \bibinfo{booktitle}{\emph{ICCV}}.
  \bibinfo{pages}{9236--9245}.
\newblock


\bibitem[\protect\citeauthoryear{Wang and Neumann}{Wang and Neumann}{2018}]%
        {DA-RGBD-SS}
\bibfield{author}{\bibinfo{person}{Weiyue Wang} {and} \bibinfo{person}{Ulrich
  Neumann}.} \bibinfo{year}{2018}\natexlab{}.
\newblock \showarticletitle{Depth-aware cnn for rgb-d segmentation}. In
  \bibinfo{booktitle}{\emph{ECCV}}. \bibinfo{pages}{135--150}.
\newblock


\bibitem[\protect\citeauthoryear{Wang, Shen, and Porikli}{Wang
  et~al\mbox{.}}{2015}]%
        {SAGE}
\bibfield{author}{\bibinfo{person}{Wenguan Wang}, \bibinfo{person}{Jianbing
  Shen}, {and} \bibinfo{person}{Fatih Porikli}.}
  \bibinfo{year}{2015}\natexlab{}.
\newblock \showarticletitle{Saliency-aware geodesic video object segmentation}.
  In \bibinfo{booktitle}{\emph{CVPR}}. \bibinfo{pages}{3395--3402}.
\newblock


\bibitem[\protect\citeauthoryear{Wang, Song, Zhao, Shen, Zhao, Hoi, and
  Ling}{Wang et~al\mbox{.}}{2019b}]%
        {AGS}
\bibfield{author}{\bibinfo{person}{Wenguan Wang}, \bibinfo{person}{Hongmei
  Song}, \bibinfo{person}{Shuyang Zhao}, \bibinfo{person}{Jianbing Shen},
  \bibinfo{person}{Sanyuan Zhao}, \bibinfo{person}{Steven~CH Hoi}, {and}
  \bibinfo{person}{Haibin Ling}.} \bibinfo{year}{2019}\natexlab{b}.
\newblock \showarticletitle{Learning unsupervised video object segmentation
  through visual attention}. In \bibinfo{booktitle}{\emph{CVPR}}.
  \bibinfo{pages}{3064--3074}.
\newblock


\bibitem[\protect\citeauthoryear{Wang, Simoncelli, and Bovik}{Wang
  et~al\mbox{.}}{2003}]%
        {SSIM}
\bibfield{author}{\bibinfo{person}{Zhou Wang}, \bibinfo{person}{Eero~P
  Simoncelli}, {and} \bibinfo{person}{Alan~C Bovik}.}
  \bibinfo{year}{2003}\natexlab{}.
\newblock \showarticletitle{Multiscale structural similarity for image quality
  assessment}. In \bibinfo{booktitle}{\emph{The Thrity-Seventh Asilomar
  Conference on Signals, Systems \& Computers, 2003}},
  Vol.~\bibinfo{volume}{2}. \bibinfo{pages}{1398--1402}.
\newblock


\bibitem[\protect\citeauthoryear{Yang and Ramanan}{Yang and Ramanan}{2019}]%
        {VCN}
\bibfield{author}{\bibinfo{person}{Gengshan Yang} {and} \bibinfo{person}{Deva
  Ramanan}.} \bibinfo{year}{2019}\natexlab{}.
\newblock \showarticletitle{Volumetric correspondence networks for optical
  flow}. In \bibinfo{booktitle}{\emph{Advances in neural information processing
  systems}}. \bibinfo{pages}{794--805}.
\newblock


\bibitem[\protect\citeauthoryear{Zhang, Zhang, Lin, Mech, Lu, and He}{Zhang
  et~al\mbox{.}}{2020}]%
        {WCS}
\bibfield{author}{\bibinfo{person}{Lu Zhang}, \bibinfo{person}{Jianming Zhang},
  \bibinfo{person}{Zhe Lin}, \bibinfo{person}{Radomir Mech},
  \bibinfo{person}{Huchuan Lu}, {and} \bibinfo{person}{You He}.}
  \bibinfo{year}{2020}\natexlab{}.
\newblock \showarticletitle{Unsupervised Video Object Segmentation with Joint
  Hotspot Tracking}. In \bibinfo{booktitle}{\emph{ECCV}}.
  \bibinfo{pages}{490--506}.
\newblock


\bibitem[\protect\citeauthoryear{Zhang, Wang, Lu, Wang, and Ruan}{Zhang
  et~al\mbox{.}}{2017}]%
        {Amulet}
\bibfield{author}{\bibinfo{person}{Pingping Zhang}, \bibinfo{person}{Dong
  Wang}, \bibinfo{person}{Huchuan Lu}, \bibinfo{person}{Hongyu Wang}, {and}
  \bibinfo{person}{Xiang Ruan}.} \bibinfo{year}{2017}\natexlab{}.
\newblock \showarticletitle{Amulet: Aggregating multi-level convolutional
  features for salient object detection}. In \bibinfo{booktitle}{\emph{ICCV}}.
  \bibinfo{pages}{202--211}.
\newblock


\bibitem[\protect\citeauthoryear{Zhang, Cui, Xu, Yan, Sebe, and Yang}{Zhang
  et~al\mbox{.}}{2019}]%
        {PA-RGBD-SS}
\bibfield{author}{\bibinfo{person}{Zhenyu Zhang}, \bibinfo{person}{Zhen Cui},
  \bibinfo{person}{Chunyan Xu}, \bibinfo{person}{Yan Yan},
  \bibinfo{person}{Nicu Sebe}, {and} \bibinfo{person}{Jian Yang}.}
  \bibinfo{year}{2019}\natexlab{}.
\newblock \showarticletitle{Pattern-affinitive propagation across depth,
  surface normal and semantic segmentation}. In
  \bibinfo{booktitle}{\emph{CVPR}}. \bibinfo{pages}{4106--4115}.
\newblock


\bibitem[\protect\citeauthoryear{Zhao, Zhao, Li, and Chen}{Zhao
  et~al\mbox{.}}{2020d}]%
        {ACM_RGBD_SOD_1}
\bibfield{author}{\bibinfo{person}{Jiawei Zhao}, \bibinfo{person}{Yifan Zhao},
  \bibinfo{person}{Jia Li}, {and} \bibinfo{person}{Xiaowu Chen}.}
  \bibinfo{year}{2020}\natexlab{d}.
\newblock \showarticletitle{Is depth really necessary for salient object
  detection?}. In \bibinfo{booktitle}{\emph{ACMMM}}.
  \bibinfo{pages}{1745--1754}.
\newblock


\bibitem[\protect\citeauthoryear{Zhao, Sheng, Dong, Chang, Xu,
  et~al\mbox{.}}{Zhao et~al\mbox{.}}{2020b}]%
        {MaskFlowNet}
\bibfield{author}{\bibinfo{person}{Shengyu Zhao}, \bibinfo{person}{Yilun
  Sheng}, \bibinfo{person}{Yue Dong}, \bibinfo{person}{Eric~I Chang},
  \bibinfo{person}{Yan Xu}, {et~al\mbox{.}}} \bibinfo{year}{2020}\natexlab{b}.
\newblock \showarticletitle{MaskFlownet: Asymmetric Feature Matching with
  Learnable Occlusion Mask}. In \bibinfo{booktitle}{\emph{CVPR}}.
  \bibinfo{pages}{6278--6287}.
\newblock


\bibitem[\protect\citeauthoryear{Zhao, Pang, Zhang, Lu, and Zhang}{Zhao
  et~al\mbox{.}}{2020a}]%
        {GateNet}
\bibfield{author}{\bibinfo{person}{Xiaoqi Zhao}, \bibinfo{person}{Youwei Pang},
  \bibinfo{person}{Lihe Zhang}, \bibinfo{person}{Huchuan Lu}, {and}
  \bibinfo{person}{Lei Zhang}.} \bibinfo{year}{2020}\natexlab{a}.
\newblock \showarticletitle{Suppress and balance: A simple gated network for
  salient object detection}. In \bibinfo{booktitle}{\emph{ECCV}}.
  \bibinfo{pages}{35--51}.
\newblock


\bibitem[\protect\citeauthoryear{Zhao, Zhang, Pang, Lu, and Zhang}{Zhao
  et~al\mbox{.}}{2020c}]%
        {DANet}
\bibfield{author}{\bibinfo{person}{Xiaoqi Zhao}, \bibinfo{person}{Lihe Zhang},
  \bibinfo{person}{Youwei Pang}, \bibinfo{person}{Huchuan Lu}, {and}
  \bibinfo{person}{Lei Zhang}.} \bibinfo{year}{2020}\natexlab{c}.
\newblock \showarticletitle{A single stream network for robust and real-time
  rgb-d salient object detection}. In \bibinfo{booktitle}{\emph{ECCV}}.
  \bibinfo{pages}{646--662}.
\newblock


\bibitem[\protect\citeauthoryear{Zhen, Li, Zhou, Shang, Feng, Fang, and
  Quan}{Zhen et~al\mbox{.}}{2020}]%
        {DFNet}
\bibfield{author}{\bibinfo{person}{Mingmin Zhen}, \bibinfo{person}{Shiwei Li},
  \bibinfo{person}{Lei Zhou}, \bibinfo{person}{Jiaxiang Shang},
  \bibinfo{person}{Haoan Feng}, \bibinfo{person}{Tian Fang}, {and}
  \bibinfo{person}{Long Quan}.} \bibinfo{year}{2020}\natexlab{}.
\newblock \showarticletitle{Learning Discriminative Feature with CRF for
  Unsupervised Video Object Segmentation}. In \bibinfo{booktitle}{\emph{ECCV}}.
  \bibinfo{pages}{445--462}.
\newblock


\bibitem[\protect\citeauthoryear{Zhou, Wang, Zhou, Yao, Li, and Shao}{Zhou
  et~al\mbox{.}}{2020}]%
        {MATNet}
\bibfield{author}{\bibinfo{person}{Tianfei Zhou}, \bibinfo{person}{Shunzhou
  Wang}, \bibinfo{person}{Yi Zhou}, \bibinfo{person}{Yazhou Yao},
  \bibinfo{person}{Jianwu Li}, {and} \bibinfo{person}{Ling Shao}.}
  \bibinfo{year}{2020}\natexlab{}.
\newblock \showarticletitle{Motion-Attentive Transition for Zero-Shot Video
  Object Segmentation.}. In \bibinfo{booktitle}{\emph{AAAI}}.
  \bibinfo{pages}{3}.
\newblock


\end{thebibliography}

\end{document}